\documentclass[11pt]{elsarticle}
\usepackage{amsmath,amssymb,amsfonts}
\usepackage{algorithmic}
\usepackage{graphicx}
\usepackage{textcomp}
\usepackage{xcolor}
\usepackage{hyperref}
\usepackage{color,soul}
\usepackage{cleveref}
\usepackage{subcaption}
\usepackage{float}
\usepackage{amsthm}
\usepackage{changepage}
\theoremstyle{definition}
\newtheorem{definition}{Definition}[section]

\usepackage{booktabs}

\def\*#1{\mathbf{#1}}

\begin{document}

\title{Robustness of Decentralised Learning to Nodes and Data Disruption}

\begin{frontmatter}

    
    
    \author[inst1]{Luigi Palmieri\corref{cor1}} 
    \ead{luigi.palmieri@iit.cnr.it}
    \author[inst1]{Chiara Boldrini\fnref{label1}}
    \ead{chiara.boldrini@iit.cnr.it}
    \author[inst1]{Lorenzo Valerio\fnref{label1}}
     \ead{lorenzo.valerio@iit.cnr.it}
    \author[inst1]{Andrea Passarella}
     \ead{andrea.passarella@iit.cnr.it}
    \author[inst1]{Marco Conti}
     \ead{marco.conti@iit.cnr.it}
    \author[inst2]{J\'anos Kert\'esz}
     \ead{KerteszJ@ceu.edu}
    \affiliation[inst1]{organization={CNR-IIT},
                addressline={Via G. Moruzzi, 1}, 
                city={Pisa},
                country={Italy}}
    \affiliation[inst2]{organization={CEU},
                addressline={Quellenstrasse 51}, 
                city={Vienna},
                country={Austria}}
    \cortext[cor1]{Corresponding author.}
    \fntext[label1]{C. Boldrini and L. Valerio contributed equally to this work.}

\begin{abstract}
In the active landscape of AI research, decentralised learning is gaining momentum. Decentralised learning allows individual nodes to keep data locally where they are generated and to share knowledge extracted from local data among themselves through an interactive process of collaborative refinement. This paradigm supports scenarios where data cannot leave the data owner node due to privacy or sovereignty reasons or real-time constraints imposing proximity of models to locations where inference has to be carried out. The distributed nature of decentralised learning implies significant new research challenges with respect to centralised learning. Among them, in this paper, we focus on robustness issues. Specifically, we study the effect of nodes' disruption on the collective learning process. Assuming a given percentage of ``central" nodes disappear from the network, we focus on different cases, characterised by (i) different distributions of data across nodes and (ii) different times when disruption occurs with respect to the start of the collaborative learning task. Through these configurations, we are able to show the non-trivial interplay between the properties of the network connecting nodes, the persistence of knowledge acquired collectively before disruption or lack thereof, and the effect of data availability pre- and post-disruption. Our results show that decentralised learning processes are remarkably robust to network disruption. As long as even minimum amounts of data remain available somewhere in the network, the learning process is able to recover from disruptions and achieve significant classification accuracy. This clearly varies depending on the remaining connectivity after disruption, but we show that even nodes that remain completely isolated can retain significant knowledge acquired before the disruption. 

\end{abstract}

\end{frontmatter}

\section{Introduction}
\label{sec:intro}

While centralised AI algorithms based on data centres are necessary for many AI tasks, more and more researchers are focusing on decentralised AI~\cite{DBLP:journals/csur/BellavistaFM21}. In general, in decentralised AI, data stays local at nodes generating them, local models are trained based on local data, and then an aggregation process happens whereby nodes combine their local models. Normally, this process is iterative, which lets nodes improve their performance over time after repeated rounds of aggregation and local training. Such a decentralised approach addresses scenarios where data cannot leave their location for privacy or sovereignty reasons or where models must work close to the controlled devices due to real-time constraints.

Federated learning has been the first example of ``non-centralised" learning~\cite{DBLP:journals/csur/BellavistaFM21}. As discussed in Section~\ref{sec:relwork}, Federated learning still requires a central node that controls local models' aggregation and synchronisation. More recently, researchers have focused increasingly on completely decentralised learning schemes, where nodes collaborate based on locally trained models without the need for central coordination. In this paper, we focus on this case and address specific issues related to the robustness of the decentralised learning process in case of disruptions.

The robustness of decentralised learning is a key research question to address. Specifically, centralised learning guarantees a high degree of reliability for many reasons. First, while locally centralised, data centres are quite robust against failures due to standardised resilience solutions. Moreover, as data is centralised, once the resilience of data centre devices is guaranteed, models can be trained on full datasets. All these conditions are not present in decentralised learning, which intuitively may seem much more a fragile paradigm due to the fact that any node in the network may disappear, thus resulting in potential loss of connectivity (which hinders aggregation of models across nodes), data (which may be residing on individual nodes only), or both.

In this paper, we set out to analyse these aspects of robustness. Specifically, we consider a network of collaborating nodes, each hosting a local dataset and training a local model on it. We assume that nodes are connected according to a Barabasi-Albert network model, which has been shown to reproduce many real-world networked systems, both ``artificial” (such as computer networks) and ``natural” (such as human social networks)~\cite{barabasi2013network}. We further assume that nodes exchange local models only with their direct neighbours and aggregate the received models and the local one according to a well-known averaging policy (known in the literature as Decentralised Averaging~\cite{mcmahan_communication-efficient_2017}). Finally, we assume that nodes have to solve an image classification task.

Under these assumptions, we analyse the robustness of decentralised learning under three configurations, which we denote as Case~1, Case~2, and Case~3, respectively (see Section~\ref{sec:settings} for more details), characterised by increased levels of disruptions. In Case 1, after an initial period where the decentralised process proceeds without disruptions, we assume that a certain percentage of ``central" nodes disappear. However, these nodes do not hold any local data, so the effect is a disruption of the network structure only. In Case 2, we assume that the same nodes also hold local data, and therefore, the disruption consists of the loss of both data and connectivity. Finally, in Case 3, we assume that central nodes hold a significantly higher share of data with respect to the other nodes. Therefore, Case 3 represents the loss of connectivity and a particularly extended set of data on which pre-disruption models could have been trained. In all these cases, we analyse the performance in terms of accuracy achieved by surviving nodes after a certain number of training rounds. As the considered disruptions partition the network into smaller disconnected subgraphs, we also analyse the performance inside the different connected components of connected nodes that are generated after disruption, starting from isolated nodes to the largest connected component. In all cases, we compare the accuracy achieved by the surviving nodes after the disruption in comparison to the baseline case where no disruption occurs and the case when disruption occurs at time 0. The latter represents the boundary case where surviving nodes cannot exploit any collaborative learning process before the disruption.

The main take-home messages we obtain from the results presented in this paper are the following.
\begin{itemize}
    \item First and foremost, decentralised learning proves to be remarkably robust to failures. The loss of accuracy with respect to the case when no disruption occurs is negligible in the largest connected component, and it is limited to 10 to 20\% depending on specific parameters, even for completely isolated nodes. 
    \item Knowledge persists despite disruption. In all cases and for all types of surviving nodes, the accuracy after disruption grows much larger than when disruption happens at time 0. This means that, even for isolated nodes, having been exposed to the decentralised learning process allows them to maintain a significant level of knowledge acquired from other nodes, provided this can be ``refreshed" after disruption by a small local dataset.
    \item Significant disruptions of the network structure are not particularly challenging for the learning process. Provided sufficient data is available in the overall network, nodes can achieve very high accuracy even if the most central nodes of the network disappear and the network becomes partitioned.
    \item Decentralised learning can tolerate even large loss of data. Even when disrupted nodes have much larger local datasets than the others, the latter ones are still able to compensate by jointly extracting knowledge from the data available at the surviving nodes, with a very limited reduction of the overall accuracy after a disruption.
\end{itemize}

The rest of the paper is organised as follows. Section~\ref{sec:relwork} presents the most important work related to this paper. Section~\ref{sec:system_model} describes in detail the properties of the considered network and decentralised learning process. In Section~\ref{sec:settings}, we present in detail the experimental settings. Section~\ref{sec:results} discusses the results of our simulations, while Section~\ref{sec:conclusions} concludes the paper.

\section{Related work}
\label{sec:relwork}

\subsection{Decentralised learning}

Decentralised Learning (DL) extends the typical settings of Federated Learning (FL)~\cite{mcmahan_communication-efficient_2017} by removing the existence of the central parameter server. This approach is gaining momentum as it merges the privacy benefits inherent in Federated Learning with the capabilities of decentralised and uncoordinated optimisation and learning processes. 
In~\cite{Roy2019}, a DL model is deployed for a healthcare application, enabling multiple hospitals to collaboratively train a Neural Network model while keeping their data private. Another work~\cite{savazzi_federated_2020} introduces a federated consensus algorithm that extends the FedAvg method proposed by~\cite{mcmahan_communication-efficient_2017} for use in decentralised environments, with emphasis on industrial and IoT applications. The work in~\cite{sun_decentralized_2023} proposes a Federated Decentralized Average based on SGD in their research, where they incorporate a momentum term to counterbalance potential drift caused by multiple updates, along with a quantisation strategy to reduce communication requirements. In~\cite{wink2021approach}, the authors introduce a novel method for collaborative training in DL environments by means of a secret sharing schema and a multi-party average computation, aiming to preserve privacy and security. The authors of~\cite{palmieri2024impact} explore the influence of the underlying network architecture on the learning efficacy in a fully decentralised learning system.
As discussed later in the paper, in this work we focus on DecAvg, a  learning strategy that extends the widely-used FedAvg~\cite{mcmahan_communication-efficient_2017} to the decentralized domain, and generalises similar strategies proposed in~\cite{savazzi_federated_2020,sun_decentralized_2023}. We deliberately chose not to explore more complex policies, as our primary objective is to examine the effects of network and data disruptions. Introducing more sophisticated approaches could have obscured the fundamental insights into these phenomena.

In addition to the algorithmic contributions discussed above, there have been efforts to analytically characterize decentralized learning in relation to graph topology. In~\cite{koloskova2020unified}, the authors present a comprehensive analytical framework that unifies various decentralized SGD methods. This framework supports local SGD updates, synchronous updates, and pairwise gossip updates within a network topology, and offers universal convergence rates for solving both convex and non-convex problems. Notably, their analysis operates under less stringent assumptions about network properties compared to earlier research. Similarly, in~\cite{zhu2022topology}, authors examine the generalization and stability of decentralized SGD, providing a theoretical framework that correlates the network's topology with the generalization of decentralized SGD. Their main focus is on how the spectral gap affects learning outcomes. Their analysis assumes IID data distribution and simple network topologies such as rings, grids and complete networks. Finally, \cite{badie2024initialisation} investigates the impact of network topology on the optimal model initialisation in decentralised settings, proposing a novel initialization strategy that speeds up the learning convergence in the initial phase.

\subsection{Robustness of complex networks}

As stated in the introduction, network disruption is a central concern in today's interconnected world, whether the network is a communication network, a transportation network, or a social network. On the other hand, complex networks serve as the backbone of numerous real-world systems, from biological processes and social interactions to technological infrastructures like the internet and power grids~\cite{boccaletti2006complex}. As a result, the study of complex network models undergoing network disruption has attracted a lot of attention in the field of network science. For a recent review of the issues of robustness in complex networks, we refer the interested reader to~\cite{artime_robustness_2024}.
Understanding the vulnerability and resilience of complex networks is essential for developing systems that can withstand failures and disruptions. This is particularly important in disaster management, where the goal is to minimise the impact of critical events on infrastructures and services. Different network topologies exhibit varying levels of resilience and vulnerability. For example, a centralised network is highly vulnerable to single points of failure, since all network communication flows through a central node or hub. Therefore, this topology is particularly vulnerable to disruptions that target the central node.

Within the context of complex networks and their applications to real-world scenarios, networks' structures such as the Barabasi-Albert model, are frequently used as reference models. This is because of their ability to capture the inhomogeneous connectivity distribution characteristic of many real-world networks, thanks to the scale-free property exhibited by this model.
In~\cite{albert_error_2000}, the authors explored how scale-free networks respond to random failures by examining changes in network diameter when a few nodes are removed. Their findings revealed that these networks are remarkably resilient to random failures, as the chances of a random event affecting one of the critical hubs is low due to the network’s heterogeneous topology. However, this resilience does not extend to deliberate attacks targeting the network hubs (high degree nodes). 
Random networks (such as the Erd\"os-R\'enyi graph), on the other end, are typically very robust to targeted attacks but vulnerable to random attacks, because, due their connections being more evenly distributed, the loss of any node can affect network connectivity more uniformly. 
In~\cite{holme_attack_2002}, the authors analyzed the vulnerability of scale-free networks, such as Barabasi-Albert graphs, to targeted attacks, where nodes are removed based on properties like degree or centrality. Their results showed that these networks are highly vulnerable when hubs are specifically targeted, as the removal of just a few key nodes can severely disrupt the network's overall connectivity. This highlights a significant trade-off in scale-free networks: while they handle random failures well, they are critically susceptible to intentional attacks on their most connected nodes.

\subsection{Robustness of federated learning}
In the context of FL and its robustness, in \cite{kairouz_advances_2021}, the authors provide insights into the challenges and open problems associated with FL, including network disruptions and communication failures. In FL, models are trained across multiple clients holding local data samples without exchanging them. Therefore, the FL process involves several steps, including broadcasting a model to the participating clients, local client computation, and the subsequent reporting of client updates to the central aggregator. However, as the authors highlight, this setup inherently faces various challenges that can hinder task completion due to system factors that can contribute to failures occurring at any of these steps. These failures can range from explicit communication breakdowns due to unreliable network connections to the presence of straggler clients, participants that report their outputs significantly later than their peers within the same communication round. Thus, to optimise efficiency, these straggler clients may be omitted from a communication round, even in the absence of explicit failures. The presence of stragglers is often unavoidable, which significantly impairs algorithm performance. To tackle these issues, asynchronous versions of Federated Learning have been proposed, where the central server does not wait for all the clients to complete their tasks~\cite{xie2019asynchronous,chen2020asynchronous}. However, the FL architecture is naturally prone to be vulnerable to a single point of failure, where the failure of the central server impairs the whole learning process. 

In the literature, the term ``robustness" is sometimes used to describe both resilience to failures and security threats posed by malicious actors. However, in this work, we distinguish between the two: we use ``robustness" specifically to refer to resilience against failures. The security of federated learning has received significant attention in recent years. For an in-depth overview of the current state of security and privacy challenges in FL, we refer the reader to~\cite{mothukuri_survey_2021}. The authors of this survey highlight key security concerns, particularly targeted attacks from malicious actors, such as data poisoning and model poisoning, which can undermine the integrity of FL systems.

\subsection{Robustness of decentralised learning}

While federated learning imposes a star topology, where all nodes communicate through a central server, decentralized learning removes such constraints, allowing any network topology among the nodes. As a result, decentralized learning is shaped by the structure and dynamics of the underlying complex system that governs node interactions, as well as the specifics of the learning process itself.
As DL is a relatively novel research area, both its security and its robustness remain underexplored in the literature. Regarding the topic of security in DL, authors in~\cite{feng_dart_2024} explore the models' robustness in decentralized learning against poisoning attacks, which can degrade model performance. It introduces DART, a tool for analysing DL model resilience to security threats, capable of simulating various attacks and evaluating defence mechanisms. The study compares centralized and decentralized paradigms under different attack scenarios and examines the effectiveness of defences. Authors in~\cite{feng_dart_2024} conclude that the research landscape addressing security issues in FL/DL is primarily focused on FL, with fewer studies addressing purely DL, leaving the issues surrounding system security in DL remain largely unaddressed.

From the robustness standpoint, which is the focus of this work, the closest contributions are ~\cite{ye_decentralized_2021} and~\cite{palmieri2023exploring}. The authors in~\cite{ye_decentralized_2021} took into account decentralized FL where machine learning models are trained on edge devices using unreliable communication protocols, specifically UDP. They employed a modified version of traditional DSGD, which they call Soft-DSGD that updates model parameters with partially received data. The authors show that Soft-DSGD achieves the same convergence rate as standard decentralized methods while being more communication-efficient. Numerical results highlight its improved performance in scenarios with unreliable networks. However, in their setting, they account for partial communication loss and do not address the issue of client failure. 
To the best of our knowledge, the only work that focuses on node failures in DL from a topological perspective is~\cite{palmieri2023exploring}. 
This paper extends our work in~\cite{palmieri2023exploring} in several directions. Firstly, we employ a new strategy for simulating disruptions, ensuring fairer comparisons among different learning configurations by imposing disruptions once the average accuracy of the system reaches a specific target threshold. This contrasts with the previous approach, where disruptions occurred at fixed intervals. Furthermore, a novel scenario (Case 3) has been introduced, detailing situations where disrupted nodes exhibit a disproportionate proficiency in classifying specific classes of images. With this setup, we evaluate the ability of surviving nodes to retain valuable knowledge provided by disrupted nodes after disconnection.
Moreover, the paper now includes an analysis of the impact of different centrality measures for selecting nodes to be switched off, as well as a percolation analysis to characterise changes in network connectivity as nodes are removed, detailed in Section 4.2. Additionally, for Case 1 and Case 2, we now consider a more challenging (from the learning standpoint) data distribution. Finally, a comparison of the performance of isolated nodes versus the largest connected component has been added to illustrate the robustness of the decentralised learning process in extreme cases following disruption.

\section{System model}
\label{sec:system_model}

\subsection{Decentralized learning}
\label{sec:decentralized_learning}

We represent the network connecting the nodes as a graph $\mathcal{G}(\mathcal{V},\mathcal{E})$, where $\mathcal{V}$ denotes the set of nodes and $\mathcal{E}$ is the set of edges. 
The decentralised learning algorithms designed to operate on a network of nodes are typically composed of two main blocks: one for the local training of the model using local data and the other one devoted to the exchange and aggregation of the models’ updates. These operations are executed by each node, atomically,  within a single \emph{communication round}.
Each node $i \in \mathcal{V}$ has a local training dataset~$\mathcal{D}_i$ (containing tuples of features and labels $(\*x, y) \in  \mathcal{X} \times \mathcal{Y}$) and a local model $h$ defined by weights $\mathbf{w}_i$, such that $h(\mathbf{x}; \mathbf{w}_i)$ yields the prediction of label $y$ for input $\mathbf{x}$. Let us denote with $\mathcal{D} = \bigcup_i \mathcal{D}_i$ and with $\mathcal{P}$ the label distribution in $\mathcal{D}$. In this paper, we consider both the case where $\mathcal{P}_i$ (i.e., the label distribution of the local dataset on node $i$) is different from $\mathcal{P}$, as well as the case where label distributions $\mathcal{P}_i$ are IID (independent and identically distributed) across all devices (hence, $\mathcal{P}_i \sim \mathcal{P}$). We also assume homogeneous initialisation of local models across devices, i.e., at time 0, $\*w_i = \*w_j$ for any pair of nodes $i,j$. The effect of heterogenous initialisation on decentralised learning is investigated in~\cite{valerio2023coordinationfree}, while in this paper we focus on the effect of disruption on the learning process, which is an orthogonal problem.


At time 0, the model $h(\cdot; \*w_i)$ is trained on local data by minimising a target loss function $\ell$: 
\[\widetilde{\*w}_i = \mathrm{argmin}_{\*w} \sum_{k = 1}^{|\mathcal{D}_i |} \ell(y_k, h(\*x_k;\*w_i)),\] with $(\*x_k, y_k) \in \mathcal{D}_i$. 
%
At each communication round, i.e., at each step $t$, a given node $i$ receives the parameters of the local models $\widetilde{\*w}_j$ from its neighbours $j \in \mathcal{N}(i)$ in the social graph and combines them with its local model through the following aggregation policy:
\begin{equation}
    \*w_i^{(t)} \leftarrow \frac{\sum_{j \in \mathcal{N}(i)} \alpha_{ij} \widetilde{\*w}_j^{(t-1)}}{\sum_{j \in \mathcal{N}(i)} A_{ij}},
\end{equation}
where $\mathcal{N}(i)$ is the neighbourhood of node~$i$ including itself, $A_{ij}$ is the $i,j$ element of the adjacency matrix of the network, and $\alpha_{ij}$ is equal to $\frac{|\mathcal{P}_j|}{\sum_{j \in \mathcal{N}_i} | \mathcal{P}_j|}$ (and captures the relative weight of the local dataset of node $j$ in the neighbourhood of node $i$).
Afterwards, each node begins a new round of local training using the newly aggregated local model. This goes on for a certain number of communication rounds.

This strategy, which we denote as DecAvg, extends FedAvg~\cite{mcmahan_communication-efficient_2017} to a decentralised setting and is a generalisation of similar strategies proposed in~\cite{sun_decentralized_2023,savazzi_federated_2020}. Differently from the standard Federated Learning, in fully decentralised learning settings (as the ones considered in this paper), the whole process cannot rely on the coordination and supervision of a central entity. This means that, from a node point of view and apart from degenerate cases, the number of other models it can directly access to improve its local knowledge is limited to the size of its neighbourhood. However, knowledge extracted by any given node from its own data can percolate beyond its immediate neighbourhood due to successive averaging and exchanges of models across communication rounds. Moreover, the connectivity between nodes is a property of the system, and under no circumstances can it be controlled by a node. Conversely, nodes can disappear according to possible failures. 

\subsection{Network resilience and disruption strategy}
\label{sec:resilience_general}

The malfunctioning of nodes within a network can lead to a systemic collapse, where the normal functionality of the system is impaired. Of course, the amount of damage inflicted on the network depends on its characteristics and the characteristics of the nodes that exhibit malfunctioning. The ability of a network to retain its functionality despite the failures of nodes within it is called \emph{network's resilience or robustness}~\cite{barabasi2013network, Gao_2016}. Network resilience is critical in various domains, including telecommunications, transportation, and social networks.
A crucial aspect of network resilience is the structure of the underlying network. Network connectivity plays an essential role in a system’s ability to survive random failures or deliberate attacks. Therefore, understanding how networks respond to node removal or failure is essential for designing robust and reliable systems. Of course, the quicker the network breaks apart, the less robust it is. To this end, percolation theory is a widespread technique used to address questions related to network resilience. Percolation analysis is a mathematical and statistical method used to study the behaviour of interconnected systems, such as networks, under various threats, including network disruption. 
Its goal is to investigate how the removal or failure of nodes or edges affects the overall connectivity of the network. The analysis involves simulating the process of removing nodes or edges from the network based on certain criteria. Normally, the main emphasis in percolation theory is to characterise the properties of the \emph{largest connected component} (LCC) after disrupting a certain fraction of nodes, which is defined as the largest subset of nodes that remain reachable from each other via a finite-length multi-hop path. Focusing on the LCC is clearly important to characterise how global properties of a network (such as global connectivity) are impacted by disruption. However, the same methodology can also be used to understand the impact of disruptions at a local level, i.e., by characterising their effect not only on the LCC but also on the other connected components of smaller sizes, down to nodes that become isolated.

This is how we use this methodology in this paper. Specifically, as explained in~\Cref{sec:settings_disruption_analysis}, we carry out a preliminary percolation analysis to identify the percentage of nodes to be disrupted to generate an interesting mix of connected components, from large ones to isolated ones. Then, we study the evolution of decentralised learning when such a percentage of nodes is disrupted by varying the point in time when these disruptions occur (to give more or less time to the underlying decentralised learning process to work on the original network before disruption). Specifically, we characterise the evolution of the learning process in the different components that emerge after disruption to show the interplay between the quality of the model learned before disruption, the residual connectivity, and the presence of data (or lack thereof) in the surviving components.



\section{Experimental settings}
\label{sec:settings}

Below we define the experimental settings of the analyses we carry out in this work. 

\subsection{Communication network topology}
\label{sec:settings_graph}

\begin{figure}[ht]
\centering
    \begin{subfigure}{.46\textwidth}
    \centering
    \includegraphics[width=\textwidth]{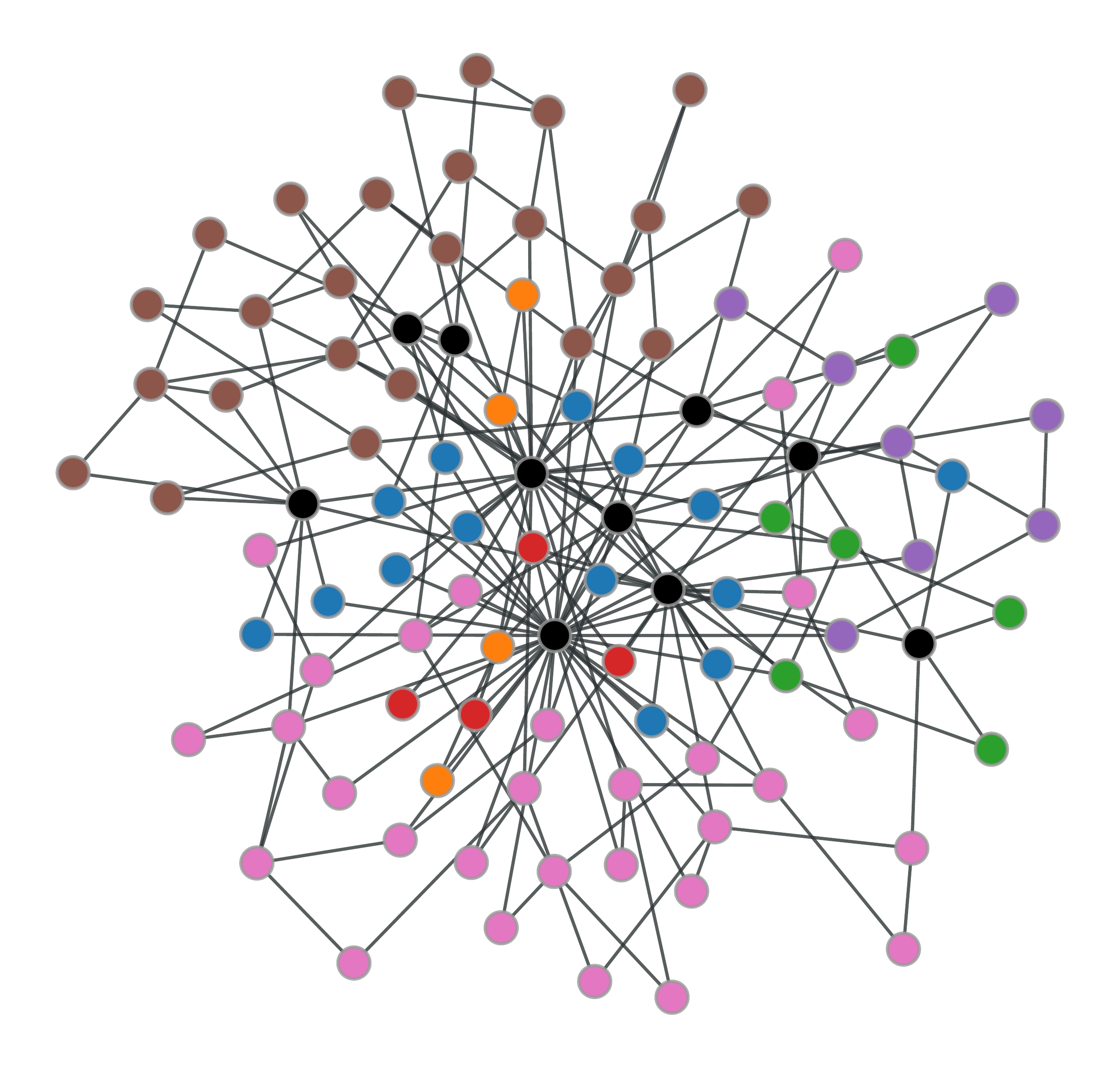}
    \caption{}
    \end{subfigure}%
    \begin{subfigure}{.45\textwidth}
    \centering
    \includegraphics[width=\textwidth]{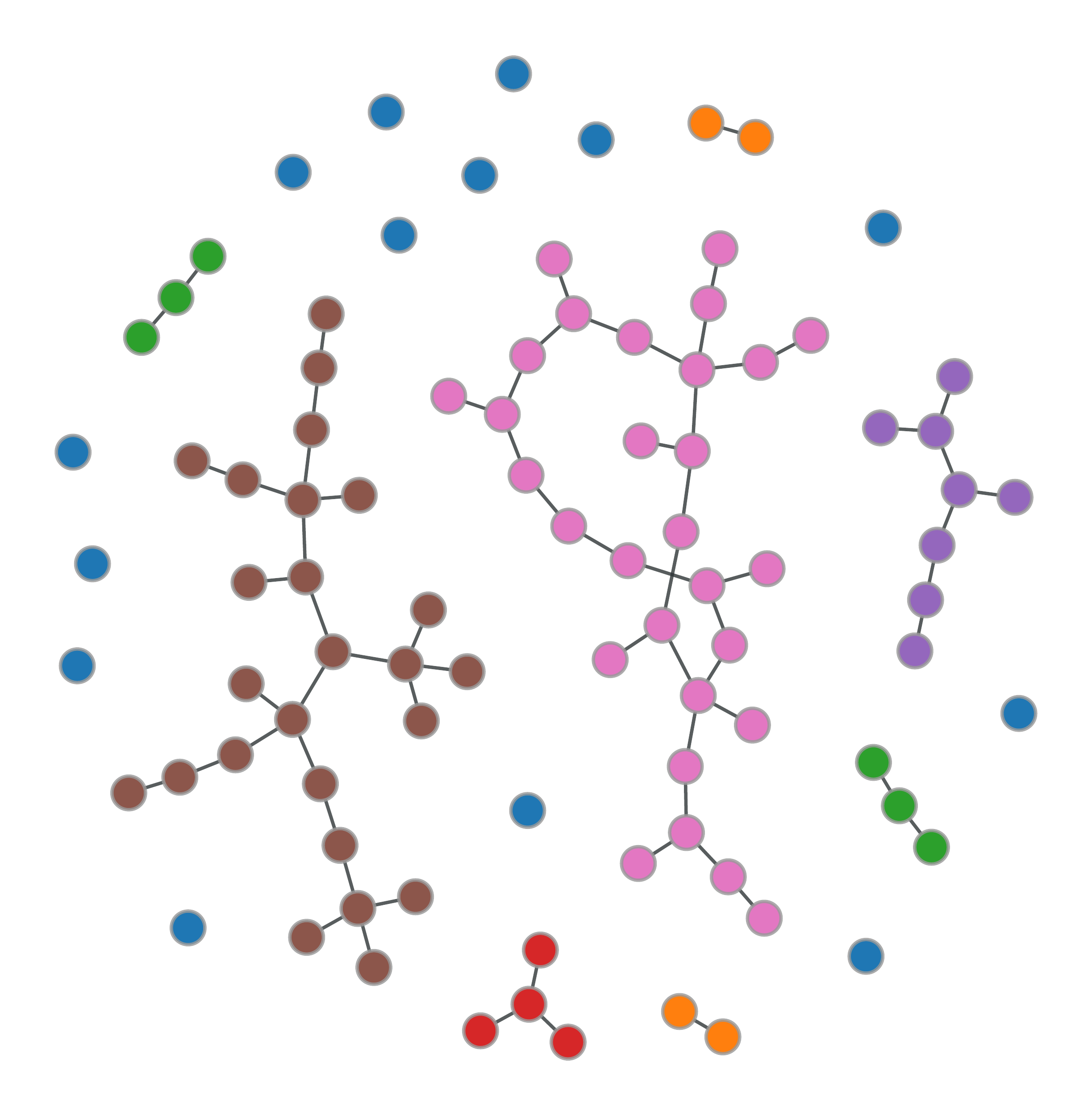}
    \caption{}
    \end{subfigure}
\caption{Barabasi-Albert Network before (left) and after (right) the disruption. The colouring of nodes is based on the size of the connected component they belong to after  disruption. The black nodes in the left image are the nodes that will be switched off when disruption takes place.}
\label{fig:net_bef_aft}
\end{figure}

We consider an unweighted Barabasi-Albert (BA) graph $\mathcal{G}$ with 100 nodes (shown in Figure~\ref{fig:net_bef_aft}) and a preferential attachment parameter $m = 2$. A BA graph is created by initialising it with $m_0 \geq m$ nodes, and then, at each step, adding a new node $i$ and connecting it with other $m$ existing nodes $j$ with a linking probability that is proportional to the degree of $j$. Since each newly created node starts off with $m$ links only, $m$ effectively becomes the minimum degree for nodes in the network. We rely on the \texttt{networkx} python library to generate this graph.

The topology of the Barabasi-Albert graph is considered able to capture important features of real-life networks by incorporating two fundamental principles: preferential attachment (new nodes in the network tend to connect to existing nodes that already have a high number of connections) and organic growth (which aligns with the dynamic nature of many real-world systems). By reproducing these principles, a BA graph model successfully emulates the power-law degree distribution observed in numerous networks, such as social networks, the World Wide Web, and collaboration networks. For this reason, it makes sense to consider it a realistic synthetic topology for our network graph. 

The analysis carried out in this paper has also been performed on the much-studied ``Zachary's Karate Club'' graph~\cite{zachary1977information}. It is a real-life network of 34 nodes corresponding to members of a sports club with links representing social relationships. The network is known to have a two-community structure documented by the separation of the club into two parts after a conflict between two instructors in the club. We selected this graph because it is based on real data and because it features a community structure that is absent in BA topologies. The results obtained for the Karate Club graph well align to the ones obtained for the BA network: it seems that the removal of highly central nodes destroys the community structure of the Karate Club graph, so that it does not have an impact on the learning process. For this reason, we decided to leave the discussion of the results related to this topology to~\ref{app:karate_club} and keep the paper focused on BA.


\subsection{Modelling disruption}
\label{sec: modelling_disruption}
In this section, we discuss how we select (i) the nodes to be disconnected from the network and (ii) the time when disruption occurs.

\subsubsection{Selection of ``disrupted'' nodes}
\label{sec:settings_disruption_analysis}

We mimic disruptions by switching off some nodes in the network while the cooperative learning task is ongoing. To maximise the effect of the disruption, these nodes are chosen based on their centrality score. It is well-known that BA graphs are sensitive to targeted attacks towards central nodes, as discussed in Section~\ref{sec:relwork}. Thus, starting with graph $\mathcal{G} = (\mathcal{V}, \mathcal{E})$, where $\mathcal{V}$ denotes the set of vertices and $\mathcal{E}$ the set of edges, we split $\mathcal{V}$ into $\mathcal{V} = \mathcal{V}_d \cup \mathcal{V}_s$, where $\mathcal{V}_d$ contains the nodes with the highest centrality score. The graph left after the disruption is $\mathcal{G}_s = (\mathcal{V}_s, \mathcal{E}_s)$, where $\mathcal{E}_s = \{ (i,j) : i,j \in \mathcal{V}_s \wedge (i,j) \in \mathcal{E} \}$.



To identify a suitable centrality metric for our study, we carry out an initial percolation analysis of the network. A percolation analysis involves monitoring the network's connectivity changes while undergoing node removal. This can be done by observing the size of the largest connected component and comparing it to the size of the original network, as follows. Let us denote the network under study with $\mathcal{G}$, composed of $N = | \mathcal{V} |$ nodes and $E = |\mathcal{E}|$ links. At time $t$, we remove $m$ nodes according to a predefined criterion. After the removal, we count the number of connected components that survive and take the one with the highest number of nodes as the largest connected component, $\mathcal{G}_c$. Then we calculate the ratio between the size of $\mathcal{G}_c$ (denoted as $N_{G_c}$) and that of the original graph $\mathcal{G}$ (denoted with $N_G$):
\[\Phi = \frac{N_{G_c}}{N_G}.\] 
The smaller $\Phi$, the higher the network fragmentation and the higher the damage the disruption has produced in the network.

\begin{figure}[t!]
\begin{subfigure}[t]{.45\linewidth}
    \centering
    \includegraphics[width =\linewidth]{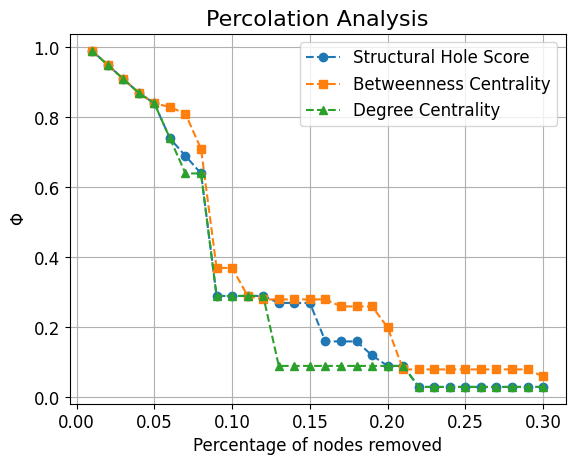}
    \caption{}
    \label{fig:perc_ratio_specific}
\end{subfigure} \hfill %
\begin{subfigure}[t]{.5\linewidth}
\centering
    \includegraphics[width = \linewidth]{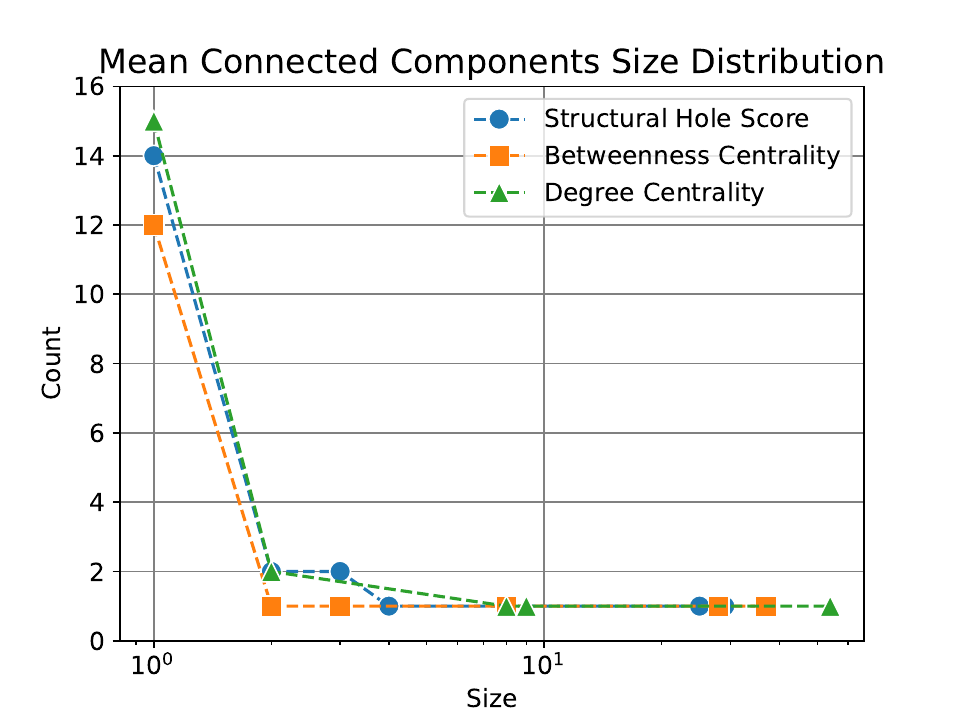}
    \caption{}
    \label{fig:connected_distribution}
\end{subfigure}\\[1ex]%
\begin{subfigure}[t]{\linewidth}
\centering
    \includegraphics[width = .45 \linewidth]{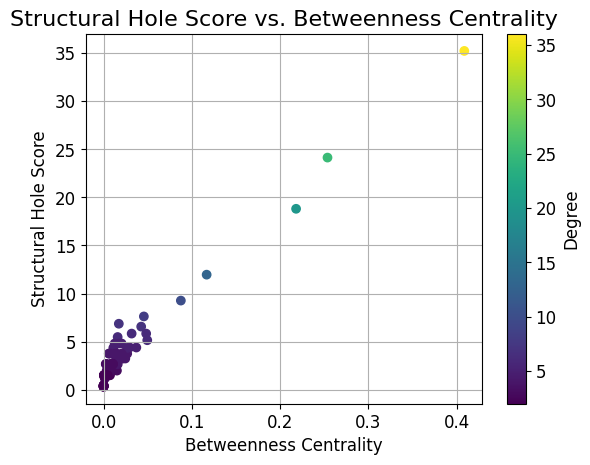}
    \caption{}
    \label{fig:scatterplot_centralities}
\end{subfigure}
\caption{Disruption analysis. (a) Size of the largest connected component as nodes are removed progressively. (b) Distribution of the size of the connected components when the top 10\% central nodes are removed. (c) Scatterplot displaying the relationship among various centrality measures. Each point (representing a vertex in the graph) is coloured according to its degree, while its x-y coordinates denote its betweenness centrality and structural hole score, respectively.}
\end{figure}

In \Cref{fig:perc_ratio_specific}, we show the behaviour of the ratio $\Phi$ in the BA network for different centrality measures: structural hole score~\cite{burt2004structural}, betweenness centrality and degree centrality. Structural holes refer to the gaps or ``holes" in a network where there are no direct connections between two or more groups of people. These gaps represent opportunities for nodes to act as intermediaries or brokers by connecting otherwise disconnected groups and facilitating the flow of information, resources, or ideas between them. Thus, the structural hole score quantifies the importance of a node in promoting information flow within the network. Nodes with higher structural hole scores are crucial in facilitating communication across different groups. The betweenness centrality quantifies the extent to which a node lies on the shortest paths between pairs of other nodes in the network. Hence, it measures the importance of a node in connecting other nodes. The degree centrality measures the importance of a node within a network based on the number of connections it has to other nodes. We refer the interested reader to~\cite{barabasi2013network} for a formal definition of these centrality metrics.

We progressively remove those nodes with the highest value of the selected centrality measure. This is effectively equivalent to analysing the robustness of the Barabasi-Albert graph undergoing targeted attacks towards the most central nodes.
As we can see in~\Cref{fig:perc_ratio_specific}, for the first 5\% of nodes removed (exactly 5 nodes in our case), the impact of all centrality measures is equal. After removing the initial 5\% of nodes, the degree centrality and structural hole score behave similarly, whereas the betweenness centrality curve remains generally higher. This implies that a larger portion of the network survives w.r.t. the other centralities, meaning that the graph is more robust to node removal when a targeted attack considers betweenness centrality.
The reason why the curves behave the same for the first 5 nodes is that their centrality values rank the same nodes regardless of the specific metric considered, as it can be seen in \Cref{fig:scatterplot_centralities}. In general, the figure shows that 
all the centrality measures have a linear directly proportional relationship. This means that nodes with high degrees also exhibit high betweenness centrality and high structural hole score. While the relationship remains directly proportional, subtle variations appear beyond the top five nodes. 
%
Other important factors in a disruption analysis are the number and size of the connected components that survive after the disruption. As we can see in \Cref{fig:connected_distribution}, after the disruption, there will be many isolated nodes (a component of size 1) and few connected components: the larger the size, the smaller the number. Note that the effect of disruption under degree centrality and structural hole score results in a higher number of isolated nodes than under the betweenness centrality case, meaning that undergoing such disruption is more detrimental to the network as it isolates more nodes.

Figure~\ref{fig:perc_ratio_specific} shows that the biggest drop in the largest connected component size happens at around 10\% node removal, with a similar pattern for all centrality measures. Therefore, we decided to remove the 10\% most central nodes. Due to its importance in measuring the importance of a node in promoting information flow, we select the structural hole score as our centrality metric. Note, though, that we are effectively also removing exactly the 10\% of nodes with the highest degree centrality, as seen in~\Cref{fig:perc_ratio_specific}.

In this work, we specifically address the scenario where nodes fail, along with all their associated edges. Failures could also be analyzed at the level of individual edges rather than entire nodes. The impact of single-edge failures on network substructures would vary based on the criteria used for edge selection, potentially leading to diverse outcomes in terms of network performance and functionality. By not focusing on individual edge removals, we concentrate on assessing the network's robustness when confronted with the complete loss of connections from a critical node. This deliberate choice allows us to explore a severe form of network impairment, providing deeper insights into the resilience of decentralized learning under extreme conditions. We believe this approach offers a compelling analysis for initial robustness studies, which can later be refined, as suggested in Section~\ref{sec:conclusions}, with more varied and nuanced failure dynamics.

\subsubsection{Time of disruption}
\label{sec:settings_timedisruption}

In disaster scenarios, disruptions present themselves in unpredictable ways. They can happen as soon as the collaborative learning process starts, in the middle, or when it is approaching its end. The loss of nodes is expected to have a different impact depending on the current progress of the learning process: weak when it happens in the final phase, stronger early on. In the latter case, the key question is whether the decentralised learning process is able to recover and catch up over time. As discussed in Section~\ref{sec:c1c2_baseline}, different learning configurations might exhibit different temporal dynamics, with one being slower than another, even if their accuracy converges to the same value. To account for these different learning speeds and enable fair comparisons among the various scenarios considered, we thus force disruptions to happen once the average accuracy of the system has reached a certain target threshold. For example, in Section~\ref{sec:c1c2_timedisruption}, we consider disruptions happening when the average accuracy in the system has reached 0.7, 0.75, 0.8. This guarantees that disruptions affect similarly ``skilled'' systems in the learning process at the time of disruption.

\subsection{Network and data resilience scenarios}
\label{sec:settings_scenarios}

\begin{figure}[t]
    \centering
    \includegraphics[width=0.3\linewidth]{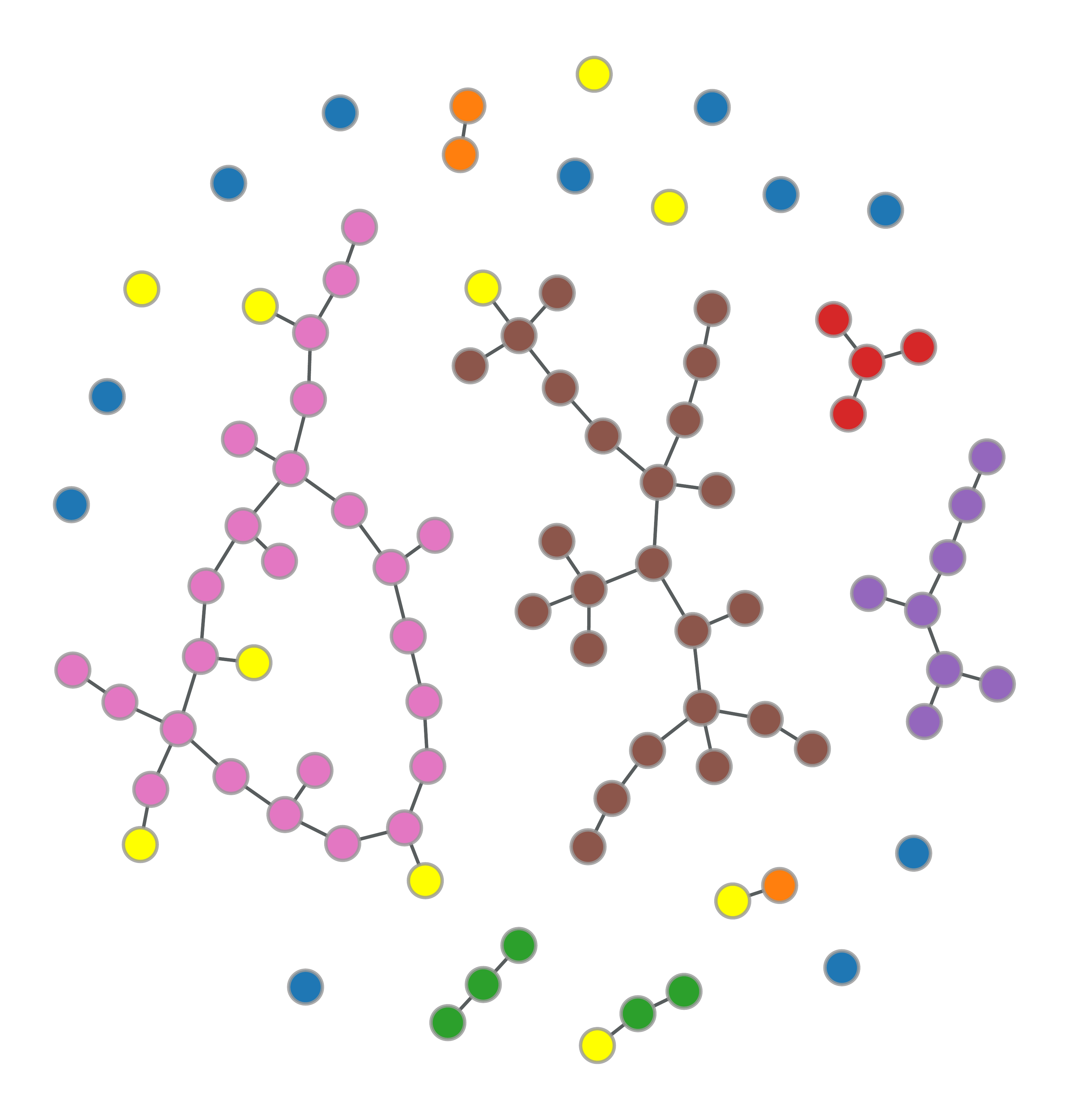}
    \caption{The figure shows the BA network after  disruption. The G2 nodes of Case 3.2 are highlighted in yellow. The rest of the coloring is size-related as before.}
    \label{fig:c3_highlight}
\end{figure}

A node in the network can contribute to the decentralised learning process in two ways. It can be well positioned in the network facilitating knowledge (i.e., model) circulation and/or can be endowed with training data to be used for model training and update. Thus, a disruption might have a different effect depending on whether the switched-off nodes only facilitate model dissemination or also possess local training data on which the local model can be trained.
For this reason, we will consider different scenarios: one in which disrupted nodes are not assigned data and two in which they are. In the latter, the highly central nodes have local data, train locally on their own data set, and share their updates with their neighbours. Thus, they are no longer just passive elements of the network. In these settings, the characteristics of the data items assigned to the disrupted nodes may play a crucial role in the resulting learning performance. For this reason, we test different data assignments to disrupted nodes.
\paragraph{Case 1: Disrupted nodes are not assigned local data}
When the disrupted nodes are not assigned local data, the highly central nodes (those cut off) will act as bridges but not contribute knowledge to the learning process. They will pass along aggregate models without using their local data for training because they don't have any. 
\paragraph{Case 2: Disrupted nodes are assigned IID local data}
In this configuration, disrupted nodes are assigned data in an IID fashion with respect to surviving nodes. In practice, this means that we assign exactly the same number of data samples per class to all nodes, regardless of whether they survive or don't survive the disruption. 
\paragraph{Case 3: Disrupted nodes are assigned non-IID local data}
In Case 3, the available class labels are split into two groups, $\mathcal{L}_1$ and $\mathcal{L}_2$. Data in $\mathcal{L}_1$ are permanently assigned in an IID fashion, meaning that all nodes are assigned the same number of samples per $\mathcal{L}_1$-class. Vice versa, $\mathcal{L}_2$-data are assigned in an unbalanced way: a group of nodes (G1) receives a low number of samples for $\mathcal{L}_2$-data, while the remaining nodes (G2) receive a high amount of $\mathcal{L}_2$-data. As a result, all nodes see an IID distribution for $\mathcal{L}_1$ data, while G1 nodes see much less $\mathcal{L}_2$-data than G2 nodes. 
Case 3 is further divided into two sub-scenarios: Case 3.1 and Case 3.2. In Case 3.1, group G2 is composed of the disrupted nodes: this means that a disproportionately high number of $\mathcal{L}_2$-data is assigned to nodes that will disappear after disruption. With this configuration, we are testing what happens when disrupted nodes contribute much more knowledge to the training process than the other nodes. This is because larger local datasets can lead to better-trained models, thus enabling more knowledge extraction. In the second scenario (Case 3.2), we instead give disproportionately more data in the $\mathcal{L}_2$-label group to the 10 nodes with the lowest structural hole score (we denote them with $\mathcal{V}_p \subset \mathcal{V}_s$, i.e., the group of non-disrupted nodes at the ``periphery'' of the network). Group G2, in this case, is composed of the nodes in $\mathcal{V}_p$. This configuration is intended to simulate a situation where the majority of the $\mathcal{L}_2$ data is located at the ``edge" of the network, reflecting scenarios where data-rich nodes are not central to the overall network structure. The key difference between the two sub-scenarios lies in what happens after the disruption: in Case 3.2, some of the G2 nodes remain connected to the network after the disruption, unlike in Case 3.1 where G2 nodes are fully isolated. This distinction is illustrated in~\autoref{fig:c3_highlight}.

\subsection{Other settings}
\label{sec:settings_other}



\paragraph{Dataset and data distribution} For our experiments, we employ the widely used MNIST image dataset~\cite{lecun1998mnist}. This dataset contains a set of handwritten digits; thus, data are divided into 10 classes (for digits from 0 to 9). The collaborative task undertaken by the nodes is then a standard image classification problem. This problem is relevant to a disaster scenario, where individuals within the affected community can leverage their smartphones or other devices with built-in cameras to capture images of the affected areas, including collapsed buildings, road blockages, or other hazardous conditions. The MNIST dataset contains 60,000 training images (approximately 6,000 per class) and 10,000 test images (1,000 per class). We evaluate the impact of the disconnection of central nodes by measuring the accuracy obtained by the local models on the standard MNIST test dataset over time. Note that all classes are equally represented in the test set and that the test set is common to all nodes.

Classification on the MNIST dataset is a relatively easy learning task. Thus, in order to stress test our decentralised learning process and to capture a realistic scenario\footnote{We anticipate that in scenarios where nodes represent user devices collaborating on decentralised learning tasks, the local datasets will be relatively small, as users actively collect data samples themselves.}, we assign very few images per class to each node. In Case 1, we distribute 7 images per class to non-disrupted nodes, leading to a local training dataset of 70 images and 6,300 images overall in the network. Recall, in fact, that in Case 1 the 10 disrupted nodes do not own any local data, so the remaining 90 nodes are the only ones assigned data. 
In Case 2, we keep the same IID data distribution, but this time, we involve disrupted nodes. In order to get approximately the same dataset size distributed across the whole network, this time, we assign 6 images per class to each node, leading to a local training dataset of 60 images and, globally, 6,000 images in the network (given that images cannot be fractional, this is the best trade-off to make Case 1 and Case 2 comparable).  
In Case 3, recall that class labels are split into two sets $\mathcal{L}_1$ and $\mathcal{L}_2$. Images belonging to $\mathcal{L}_1$ are distributed in an IID fashion across all nodes by simply splitting the approximately 6,000 images among all nodes (hence, each node gets around 60 images per class). 
In contrast, the images from $\mathcal{L}_2$ are distributed unevenly to emphasize specific differences between sub-scenarios. A small number of $\mathcal{L}_2$ images (configurations of 10, 20, or 30 images per class) are assigned to a selected group of nodes in G1, while the remaining images are allocated to G2 nodes. In both sub-scenarios, the G2 nodes are ten, thus resulting in each of these nodes having around 500 images per $\mathcal{L}_2$ class. 
%

The three different scenarios and the corresponding data distributions are summarised in Table~\ref{tab:summary_simulations}. 

\begin{table}[t]
\centering
\scriptsize
\begin{tabular}{@{}lll@{}} 
    \toprule
     & \textbf{Disrupted nodes} & \textbf{Surviving nodes} \\ 
    \midrule
    \textbf{Case 1} & No data & All classes, 7 images per class \\ 
    \textbf{Case 2} & All classes, 6 images per class & All classes, 6 images per class \\ 
    \textbf{Case 3.1} & IID in $\mathcal{L}_1$, high $\mathcal{L}_2$& IID (60 images/class) in $\mathcal{L}_1$, low $\mathcal{L}_2$\\
    \textbf{Case 3.2} & IID (60 images/class) in $\mathcal{L}_1$, low $\mathcal{L}_2$ & \begin{tabular}[t]{@{}l@{}}Low SHS: IID (60 images/class) in $\mathcal{L}_1$, high $\mathcal{L}_2$\\ Others: IID (60 images/class) in $\mathcal{L}_1$, low $\mathcal{L}_2$ \end{tabular}  \\
    \bottomrule
\end{tabular}
\caption{Summary table for the data distributions in the considered scenarios. Low $\mathcal{L}_2$-data is tested for $\{10,20,30\}$ images/$\mathcal{L}_2$-class. The remaining images per $\mathcal{L}_2$-class are split among the nodes receiving a high number of $\mathcal{L}_2$-data. Acronym SHS stands for Structural Hole Score.}
\label{tab:summary_simulations}
\end{table}

\paragraph{Decentralised learning strategy and local model architecture}
We use the DecAvg scheme implementation within the SAISim simulator, available on Zenodo~\cite{lorenzo_valerio_2021_5780042}. The simulator is developed in Python and leverages state-of-the-art libraries such as \texttt{pytorch} and \texttt{networkx} for deep learning and complex networks, respectively. It also implements primitives to support fully decentralised learning. The decentralised learning process is run for 200 communication rounds.
For the learning task, we consider a simple classifier as the learned model. The local models of nodes are Multilayer Perceptrons with three layers (sizes 512, 256, 128) and ReLu activation functions. SGD is used for the optimisation, with a learning rate of 0.01 and momentum of 0.5.

\subsection{Evaluation metrics}
\label{sec:eval_metrics}

Here, we introduce the metrics used to evaluate the robustness of decentralised learning to node disruption. A direct measure of classification performance is the accuracy achieved by nodes' local models\footnote{Note that \emph{local} models here are the result of progressively averaging models obtained from neighbours. Thus they embed global knowledge.} on the test set, which is common to all nodes. The individual accuracy $A_i(t)$ for a node~$i$ at communication round $t$ is defined as the ratio between the number of correct predictions with the local model at $t$ on the test set and the total number of samples in the test set. Given that disrupted nodes do not participate in the learning any more after disruption, we do not consider the accuracy they achieve in our analyses. To obtain a compact view of the performance, we also consider the average accuracy across non-disrupted nodes, which is computed as follows.
\begin{definition}[Mean Accuracy]
    The mean accuracy of the system is defined as the mean over all the surviving nodes:
    \begin{equation}
        \overline{A}(t) = \sum_{i \, \in \,\mathcal{V}_s} \frac{1}{|\mathcal{V}_s|}A_i(t),
    \end{equation}
    where $\mathcal{V}_s$ denotes the set of surviving nodes and $A_i(t)$ the accuracy at communication round $t$ for node $i \in \mathcal{V}_s$.
\label{def: mean_acc_definition}
\end{definition}
\noindent $\overline{A}(t)$ will be used to measure the overall system performance. 

When the top 10\% central nodes are disrupted, the connectivity in the graph drastically decreases. As illustrated in Figure~\ref{fig:connected_distribution}, many nodes become completely disconnected, while others form smaller connected components\footnote{Recall that a connected component is defined as a subgraph where a path exists between any node pair.}. Reduced connectivity results in fewer opportunities for communication and collaboration, impairing the decentralised learning process. It is anticipated that nodes losing more connections, and particularly those becoming isolated, will be affected most severely. Thus, we also define the mean accuracy for the similarly sized connected components as follows.
\begin{definition}[Mean accuracy within same-size connected components]
    The mean accuracy over connected components of size $c$ is defined as the mean over all $\mathcal{V}_s$ nodes belonging to connected components of the chosen size $c$.
    \begin{equation}
        \overline{A}^{c}(t) = \sum_{i\,\in\, \mathcal{C}_c}\frac{1}{|\mathcal{C}_c|} A_i(t)
    \end{equation}
where $\mathcal{C}_{c}$ denotes the set of nodes belonging to a connected component of size $c$.
\label{def:mean_over_clusters}
\end{definition}
\noindent The mean accuracy over connected components will be used to measure a more fine-grained impact of the disruption on the system, accounting for localised effects. Note that we aggregate over all nodes belonging to connected components with size $c$, regardless of whether they are in the same connected component. In the case of connected components of size 1, for example, this allows us to consider all isolated nodes together. 

The metrics defined above will be used to quantify the impact of the disruption. However, in order to measure differences between the various cases under investigation, we will employ a percentage difference metric. This metric will indicate the distance in performance between the different study cases.
\begin{definition}[Accuracy Difference]\label{def:accuracy_diff}
    The accuracy difference between two study cases $i$ and $j$, be them different data distribution or different accuracy thresholds at which the disruption occurs, is defined as:
    \begin{equation}
        d_A(A_i^k,A_j^k;t) = \frac{A_{i}^k(t)}{A_{j}^k(t)} - 1
    \end{equation}
    where $A^k(t)$ is the accuracy of the $k$-th node, and the subscript refers to the accuracy measures in the different cases $i$ and $j$. To denote the average across all nodes, we will use the notation $d_A(A_i,A_j;t)$.
\end{definition}
This metric serves as a tool for evaluating and comparing various study cases and disruption conditions within our analysis. By examining the sign of the function, we can discern which study case exhibits superior accuracy, whether at a global or local level. Furthermore, the function's magnitude yields the percentage variation among the many scenarios, facilitating a comparison between their relative efficacies.
Note that, whenever we compute accuracy differences, we align communication rounds at the time of disruption. Specifically, we consider the accuracies of the two case studies $i$ and $j$ at $t$ communication rounds after their disruption events (which may happen at different communication rounds in the two cases, as exemplified in~\Cref{tab:drop_times}). This way, we are sure to have given the two study cases the same amount of time to recover from disruptions.

Please note that all average results in Section~\ref{sec:results} will be reported with their corresponding 95\% confidence intervals.  In certain instances, the interval may be so narrow that it is not readily apparent.

\section{Results}
\label{sec:results} 

In this section, we discuss the impact of central nodes being cut off from the network due to a disruption, and we focus on the three scenarios, Case~1, Case~2 and Case~3, illustrated in the previous section. Recall that in Case~1 and Case~2, nodes have an IID distribution of local data, with the difference that in Case 1, highly central nodes, i.e the ones with highest structural hole score, do not have local data, and thus they contribute only by connecting elements in the graph topology. The data that these nodes hold locally in Case~2 are redistributed in Case 1 among the other nodes, preserving the IID property among them to maintain the total number of data points equal at the overall network level. Case 3 is different from the other two, as we use it to test the impact of disruption when the data distribution across nodes is non-IID.
First, in Sections~\ref{sec:c1c2_baseline}-\ref{sec:c1c2_knowledge_persistence}, we discuss and compare Case 1 and Case 2, as in both the disrupted nodes do not enjoy more knowledge than the other nodes. Then, Case 3 will be discussed separately in Section~\ref{sec:results_case3}. Due to the significantly different data distributions between Cases 1-2 and Case 3, a direct quantitative comparison would not be appropriate.

\subsection{Baseline performance when no disruption occurs}
\label{sec:c1c2_baseline}

\begin{figure}[t]
    \centering
    \includegraphics[width=.7\textwidth]{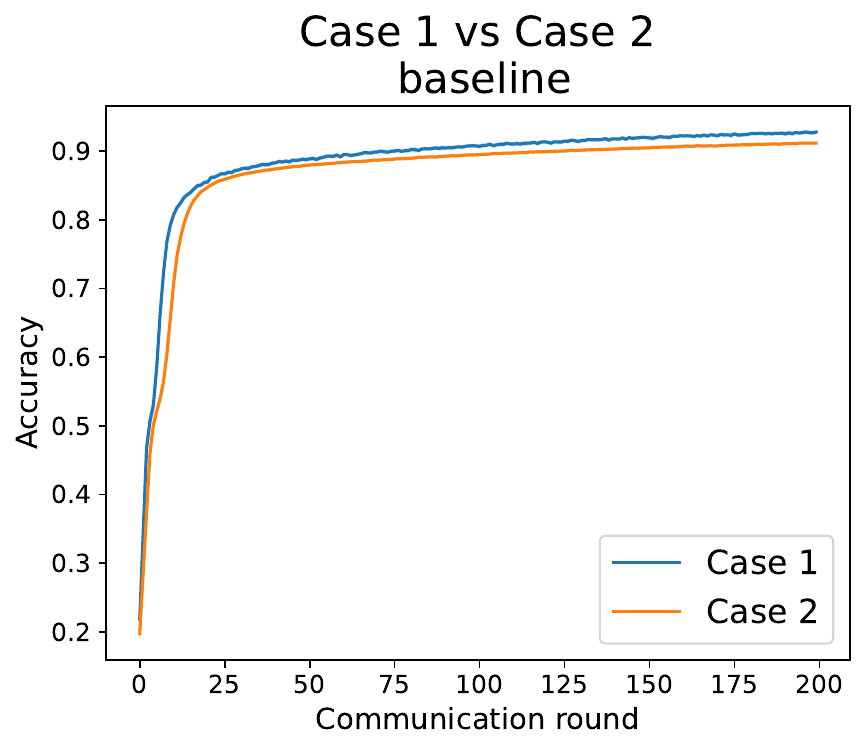}
    \caption{Mean accuracy of the system in Case 1 (blue curve) and Case 2 (orange curve).}
    \label{fig:baseline_comparison}
\end{figure}

In \Cref{fig:baseline_comparison}, we show the global system's performance, as defined in \Cref{def: mean_acc_definition}, in Case 1 and Case 2 when no disruption occurs. As evident from the plot, the curve representing Case 2 (orange curve) exhibits a sharper increase and overall higher accuracy than its counterpart in Case~1. In Case 2, the central nodes engage in local training in addition to their role as model aggregators and relays across the network. Despite a similar total dataset size between the two cases (6,300 images for Case 1 and 6,000 images for Case 2, with a slight advantage for Case 1) and a slightly larger local training dataset in Case 1, the superior accuracy achieved by Case~2 implies that bridge nodes offer more informative updates when engaged in local training. This enhancement compensates for the smaller training datasets, both locally and overall. 
With increasing communication rounds, the Case~1 curve tends to the Case 2 curve. In other words, if we allow the system adequate time for information exchange, it can compensate for any informative updates missing initially, ultimately achieving the same accuracy level as its Case 2 counterpart. This equivalence arises from the similarity in the amount of data available at the system level in both cases. From the perspective of disruption, this finding implies that if the disruption were to occur later on, the consequences of connectivity loss and the concurrent loss of connectivity and data would be identical.

\Cref{fig:baseline_comparison} showcases what we already anticipated in Section~\ref{sec:settings_timedisruption}: when two learning processes have differing speeds, a disruption occurring at a fixed time $t$ will encounter two systems at distinct stages of the learning process. While this scenario realistically captures the notion that faster systems are more likely to have acquired sufficient knowledge when disrupted, comparisons between the two systems for what-if analyses may be less informative. The discrepancy between the two scenarios is depicted in \Cref{tab:drop_times}, which illustrates the communication round at which the mean accuracy of the system reaches the specified accuracy threshold in both cases. As expected, Case 2 reaches the accuracy threshold faster than Case 1. Therefore, as discussed in Section~\ref{sec:settings_timedisruption}, in this study, we chose to assess the robustness of the two systems when disruptions occur at the same average accuracy level for both. The objective is to compare the consequences of disruption in systems that had achieved similar skill levels in terms of learning.

\begin{table}[t]
\centering
\begin{tabular}{@{}lrrr@{}}
\toprule
      & \textbf{Accuracy 0.7} & \textbf{Accuracy 0.75} & \textbf{Accuracy 0.8} \\ \midrule
\textit{Case 1} & 33           & 37            & 46           \\
\textit{Case 2} & 23           & 26            & 34           \\ \bottomrule
\end{tabular}
\caption{Communication round at which the mean accuracy of the system reaches the selected accuracy threshold, in the Case 1 and Case 2 baselines.}
\label{tab:drop_times}
\end{table}

\subsection{Impact of the time of disruption}
\label{sec:c1c2_timedisruption}

\begin{figure}[ht]
\begin{adjustwidth}{-1cm}{-1cm}
    \begin{subfigure}{1.2\textwidth}
        \includegraphics[width=\textwidth]{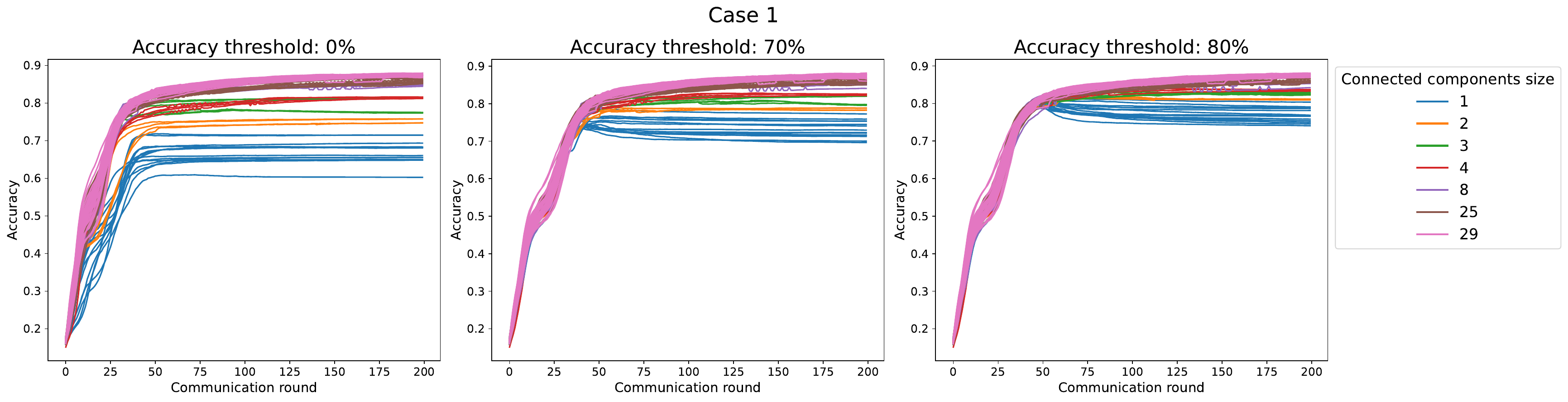}
        \caption{}
        \label{fig:case1_accuracy}
    \end{subfigure}
    
    \begin{subfigure}{1.2\textwidth}
        \includegraphics[width=\textwidth]{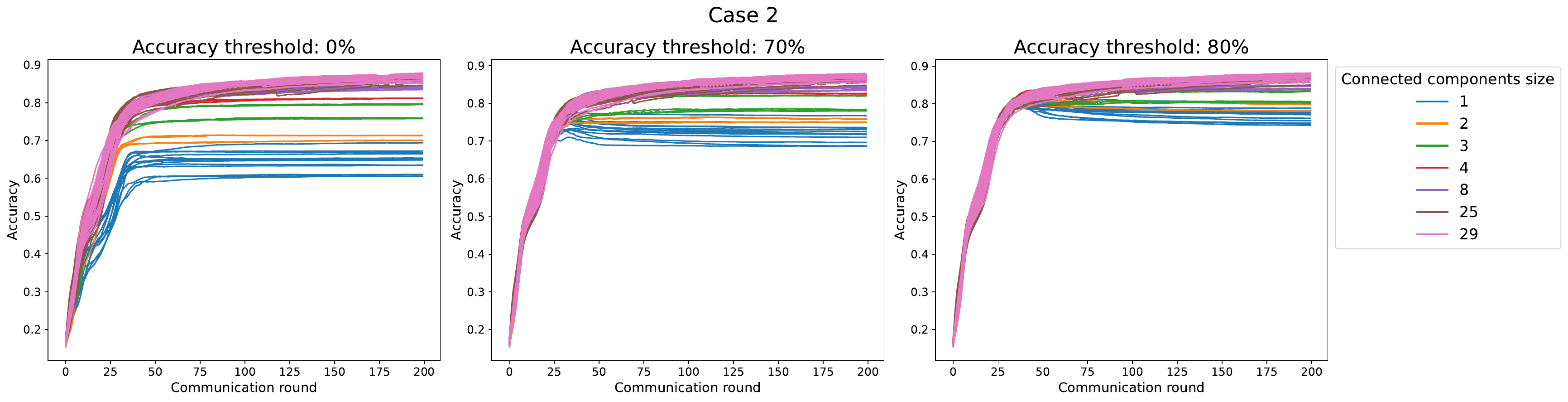}
        \caption{}
        \label{fig:case2_accuracy}
    \end{subfigure}
    \end{adjustwidth}
    \caption{Accuracy curves for each surviving node in Case 1 (a) and Case 2 (b). From left to right: increasing accuracy threshold condition for the disruption. The curves are coloured based on the size of the connected component to which the corresponding node belongs after the disruption. Accuracy threshold 75\% is omitted for ease of visualisation, as it only showed an intermediate behaviour between threshold 70\% and 80\%.}
    \label{fig:accuracy_all}
\end{figure}

We now start investigating the impact of disruptions. 
In~\autoref{fig:accuracy_all}, we show the accuracy of all the surviving (i.e., non-switched-off) nodes in Case 1 and Case 2 when the disruption happens at different accuracy thresholds. The colouring here is defined based on the size of the connected component the nodes belong to after the disruption occurs, as discussed in Section~\ref{sec:settings_disruption_analysis}.

First, we focus on the effect of different disruption times in~\autoref{fig:accuracy_all}. The later the disruption happens, i.e., going from left to right, the higher the mean accuracy of the system, as more and more nodes have been able to reach a higher level of accuracy before the cut-off, resulting in a narrower curve beam. This result holds for both scenarios and follows naturally after noticing that the central nodes are able to connect the network and, in Case 2, share their information for more communication rounds before getting switched off as the accuracy threshold increases. In all cases, after the disruption occurs, most curves stop improving and tend to cluster into different groups of similarly performing nodes. As the curve colours show, these groups reflect the connected component size to which the nodes belong. Recall that since the connected components' sizes are calculated after the removal of the special nodes, nodes having connected component size 1 are isolated nodes. As we can see in Figure~\ref{fig:accuracy_all}, isolated nodes have the lowest values of accuracies, and the accuracy generally increases with the increasing value of the connected component size. This comes naturally from the fact that more connected nodes enjoy more information flow due to collaborative learning, thus increasing their performance. Furthermore, by examining the curves for the two largest connected components (size 25 and size 29), we can observe that the larger the connected component size, the more a node's performance becomes independent of the disruption time. This indicates that the connected component has sufficient data to compensate for the disruption's effects, no matter when it occurs.
Conversely, we notice a strong correlation between the accuracy level and the timing of disruptions for the isolated nodes: the later the disruption occurs, the better the accuracy level that the isolated nodes can achieve. Naturally, this is because they have longer access to more information. Connected components of intermediate sizes fall within these two boundary conditions.

Interestingly, in~\autoref{fig:accuracy_all}, we note that for later disruption times, isolated nodes, as well as nodes belonging to small connected components, reduce the accuracy they achieve with respect to the one they had achieved at the disruption time. This effect is not present for nodes in larger connected components. The intuition is that large connected components can compensate for the lack of connectivity thanks to the overall set of data residing at the nodes in the connected component. 

\emph{Take-home:} If the connected component is large enough, the effect of disruption can be compensated reasonably well. Otherwise, nodes in smaller connected components are not able to recover knowledge that is lost as a side effect of disruption. We analyse this effect in more detail in the next section.


\subsection{Connectivity loss vs connectivity and data loss}
\label{sec:c1c2_connectivity_vs_data_loss}

\begin{figure}[t]
    \centering
    \includegraphics[width=.7\textwidth]{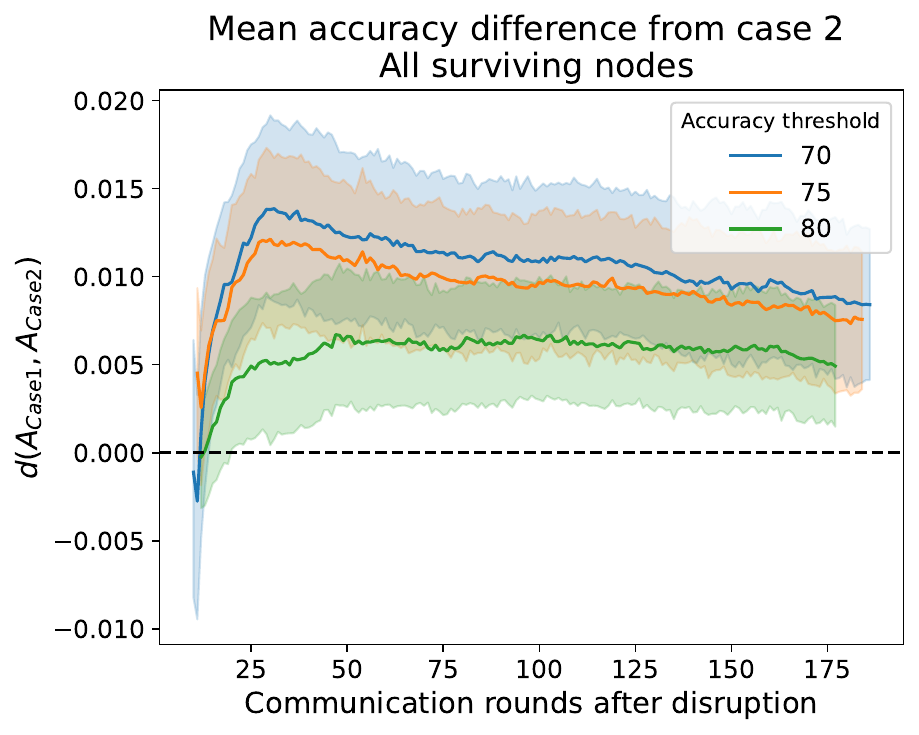}
    \caption{Average accuracy difference between Case 1 and Case 2. Communication rounds are aligned at the time of disruption for both cases.}
    \label{fig:case1_vs_case2_global}
\end{figure}

\begin{figure}[t]
\centering
    \begin{subfigure}{.5\textwidth}
    \centering
     \includegraphics[width=\textwidth]{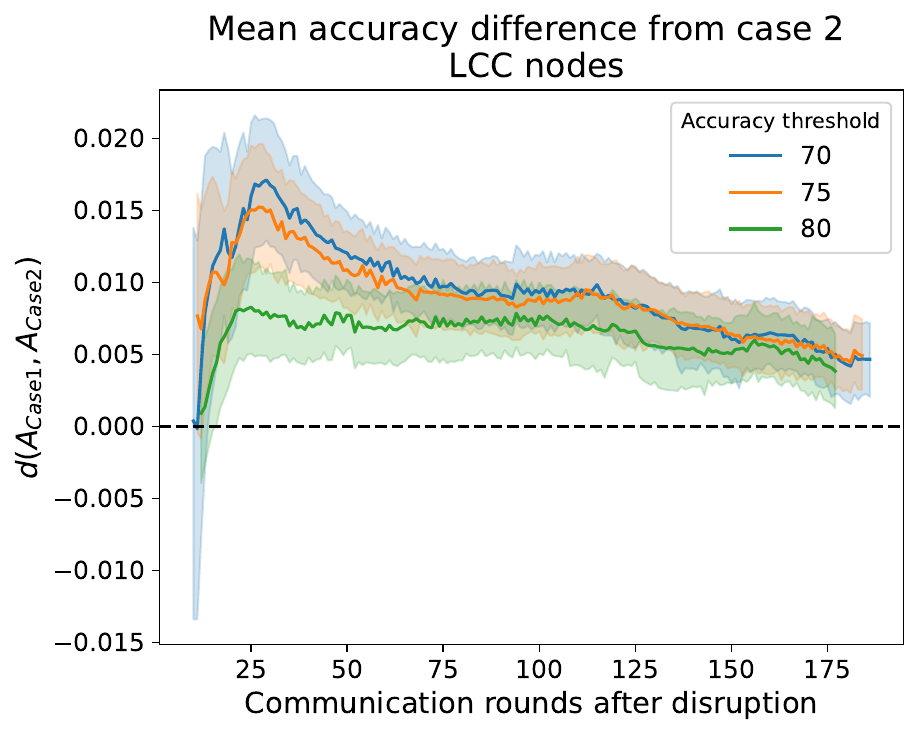}
     \caption{}
     \label{fig:case1_vs_case2_large_cluster}
    \end{subfigure}%
    \begin{subfigure}{.5\textwidth}
    \centering
    \includegraphics[width=\textwidth]{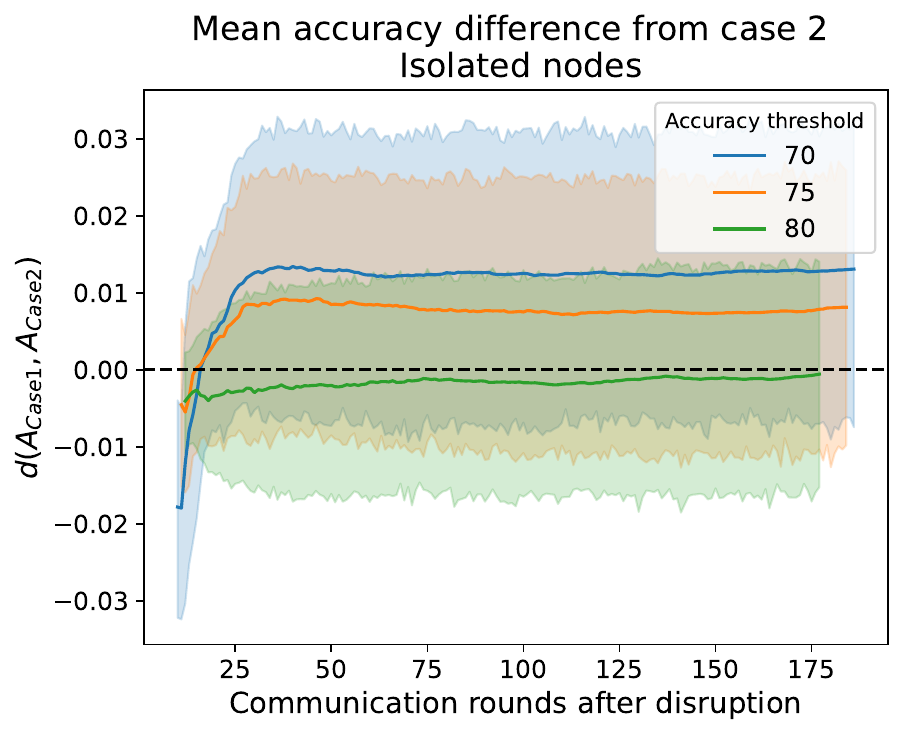}
    \caption{}
    \label{fig:case1_vs_case2_isolated_nodes}
    \end{subfigure}
    \caption{Mean local accuracy difference for the largest connected component (left) and the isolated nodes (right).}
    \label{fig:case1_vs_case2_local}  
\end{figure}

In this section, we directly compare Case 1 and Case 2 with respect to the evolution of the decentralised learning process after disruption. 
Specifically, in \Cref{fig:case1_vs_case2_global}, we show the difference between the overall performances of the two cases, denoted as $d_A(A_{Case 1}, A_{Case2})$, for the different accuracy thresholds, where $d_A(A_{Case 1}, A_{Case2})$ is the mean of the accuracy differences over all surviving nodes. The accuracy difference is computed at each communication round following the time of disruption. As we can see, the curves are all positive and stabilise at small distances. Even though, after a large number of communication rounds, the difference tends to be very limited, this indicates that, in general, the configuration of Case 1 is more beneficial than that of Case 2 to limit the damage of disruption. 

This is further confirmed at a more granular level when we compute the same difference but focusing on the largest connected component (\Cref{fig:case1_vs_case2_large_cluster}) and the isolated nodes (\Cref{fig:case1_vs_case2_isolated_nodes}), thus computing the difference between the within-component accuracy, as per~\Cref{def:mean_over_clusters}. The behaviour of the large connected component is similar to that observed for the system overall. Specifically, we note that (i)~the condition of Case 1 is more favourable than that of Case~2 and that (2) after sufficient time, the accuracy threshold at which disruption occurred does not differentiate the two conditions. Moreover, the case of isolated nodes shows that when disruption occurs at a (relatively) low accuracy, the presence of even a small amount of additional data locally allows the model of Case 1 to reach a higher accuracy than the model trained in Case 2. However, this difference vanishes when disruption occurs at a high accuracy threshold.

\emph{Take-home:} All in all, our results suggest that having more data available ``somewhere" in the network (as in Case 1) is more beneficial than concentrating the same amount of data on central nodes, if those nodes are at risk of disruption (as in Case 2), while the main advantage of the latter configuration is limited to a speed-up of the decentralised learning process if no disruption occurs (as shown by~\Cref{fig:baseline_comparison}). This confirms that, as long as data are present in the network, decentralised learning is able to exploit them through collaborative training, even in the presence of significant disruptions to the network structure.

\subsection{Knowledge persistence}
\label{sec:c1c2_knowledge_persistence}

In this section, we analyse in detail the aspect of knowledge persistence, referring to the nodes' ability to retain the memory of the knowledge transmitted by nodes that were disrupted. Specifically, we analyse, after disruption, the loss of accuracy with respect to the case where no disruption occurs. In the following, ``baseline" and ``accuracy threshold 0" refer to the cases where the disruption either never happens or has already happened at the start of the simulation, respectively. They represent the upper and lower bounds of achievable accuracy. 

\begin{figure*}[t]
    \centering
    \begin{subfigure}{\textwidth}
    \centering
    \includegraphics[width=\textwidth]{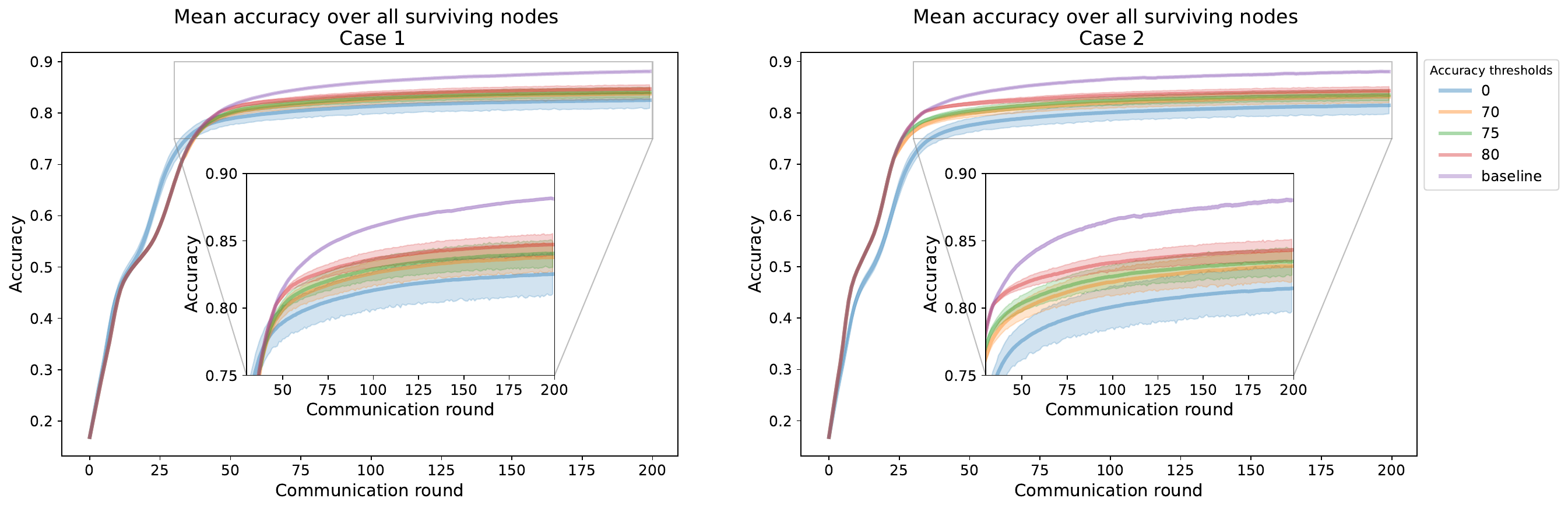}
    \caption{}
    \label{fig:c1c2_persistence_mean_acc_all}
    \end{subfigure}
    \begin{subfigure}{\textwidth}
    \centering
    \includegraphics[width=\textwidth]{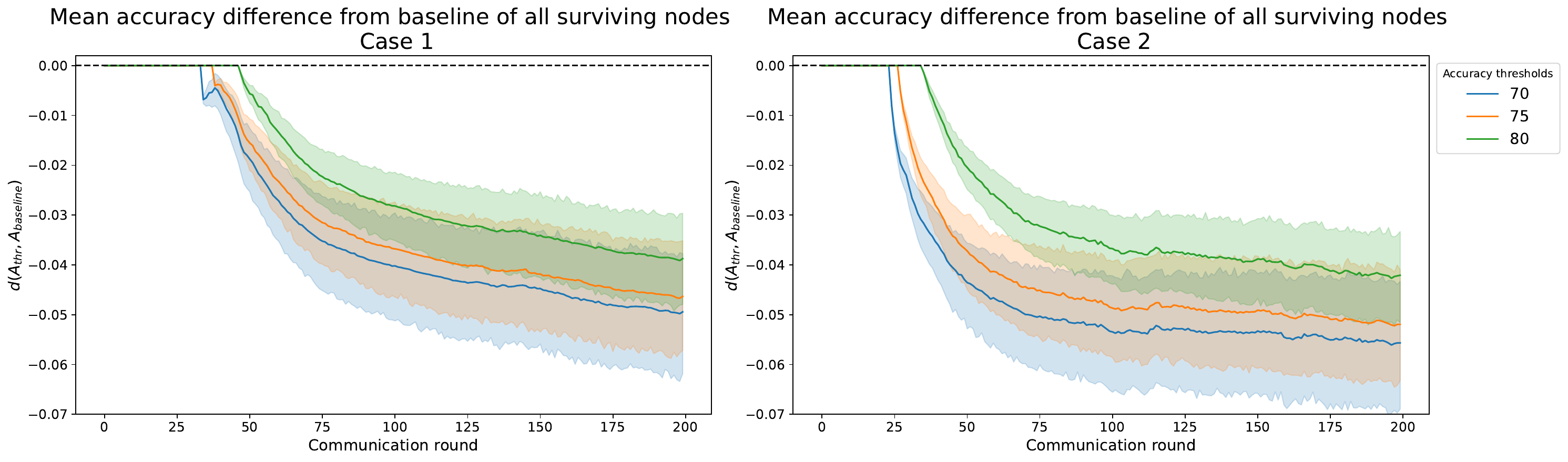}
    \caption{}
    \label{fig:c1c2_persistence_distance_baseline_all}
    \end{subfigure}
    \caption{Knowledge persistence analysis (all nodes). (a) Average accuracy of the system calculated among all non-switched-off nodes for different accuracy thresholds. (b) Distance between the different accuracy thresholds and the baseline. The distance is calculated as the percentual difference between the accuracies.}
    \label{fig:c1c2_persistence_all}
\end{figure*}

\begin{figure}[t]
    \centering
    \begin{subfigure}{\textwidth}
    \centering
    \includegraphics[width=\textwidth]{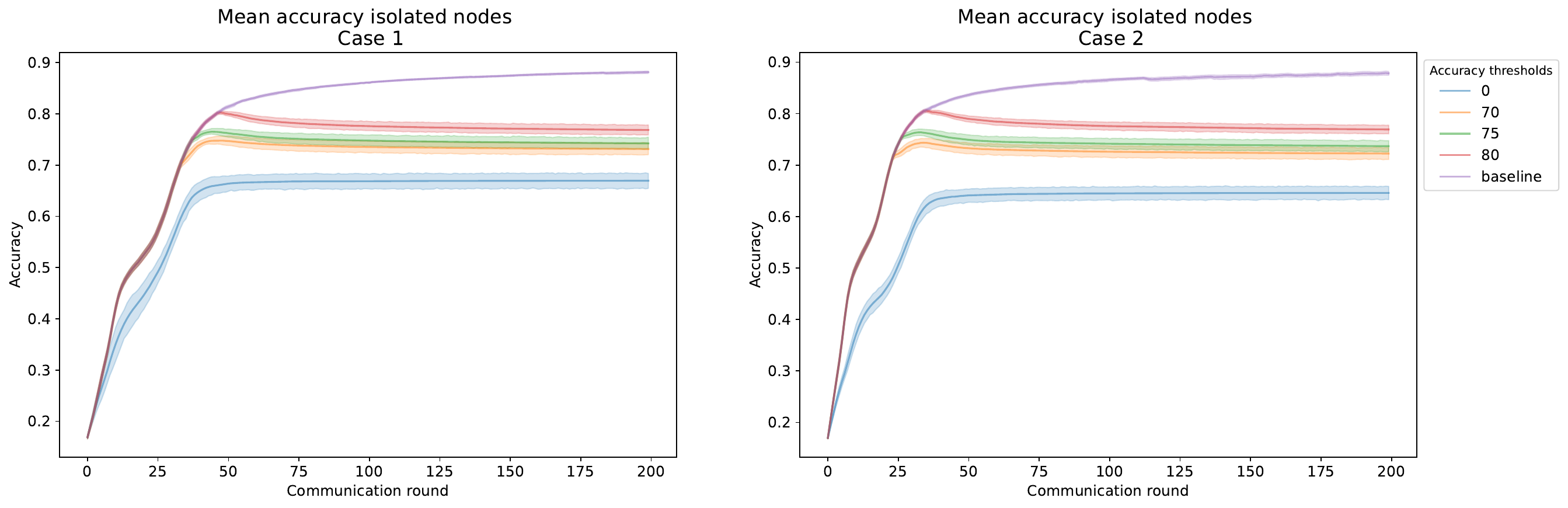}
    \caption{}
    \label{fig:c1c2_persistence_mean_acc_isolated}
    \end{subfigure}
    \begin{subfigure}{\textwidth}
    \centering
    \includegraphics[width=\textwidth]{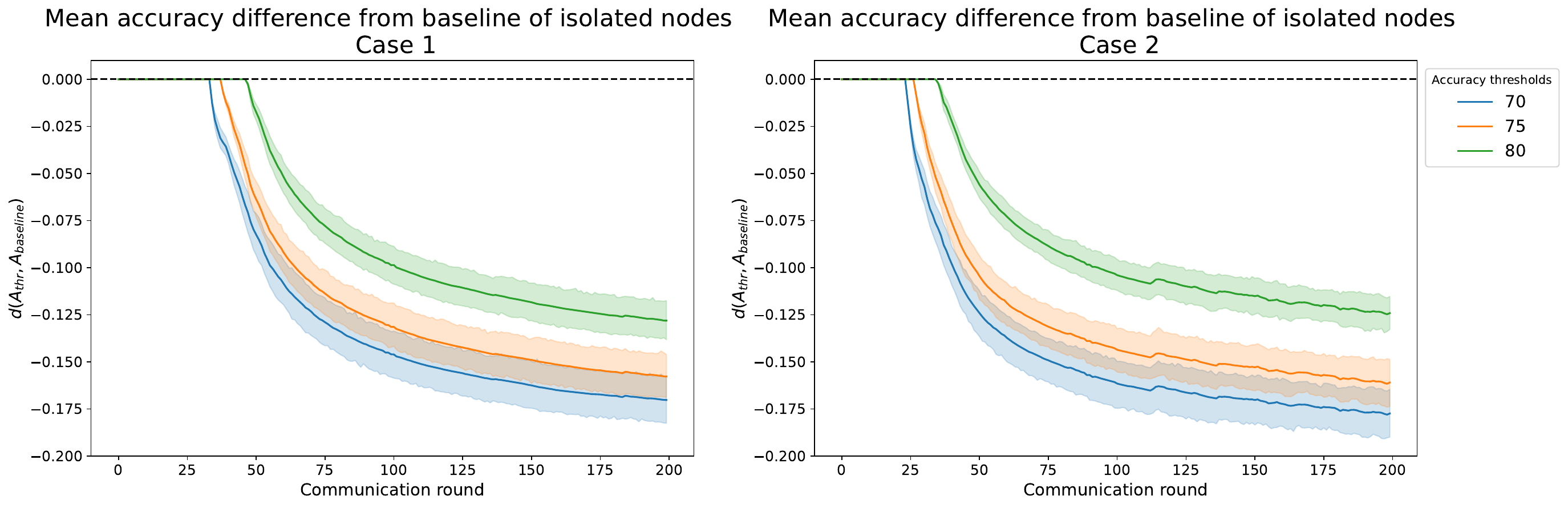}
    \caption{}
    \label{fig:c1c2_persistence_distance_baseline_isolated}
    \end{subfigure}
    \caption{Knowledge persistence analysis (isolated nodes). (a) Mean accuracy over all isolated nodes, for different disruption thresholds. (b) Mean  accuracy difference with respect to baseline over all isolated nodes, for different disruption thresholds.}
    \label{fig:c1c2_persistence_isolated} 
\end{figure}

\begin{figure}[t]
    \centering
    \begin{subfigure}{\textwidth}
    \centering
    \includegraphics[width=\textwidth]{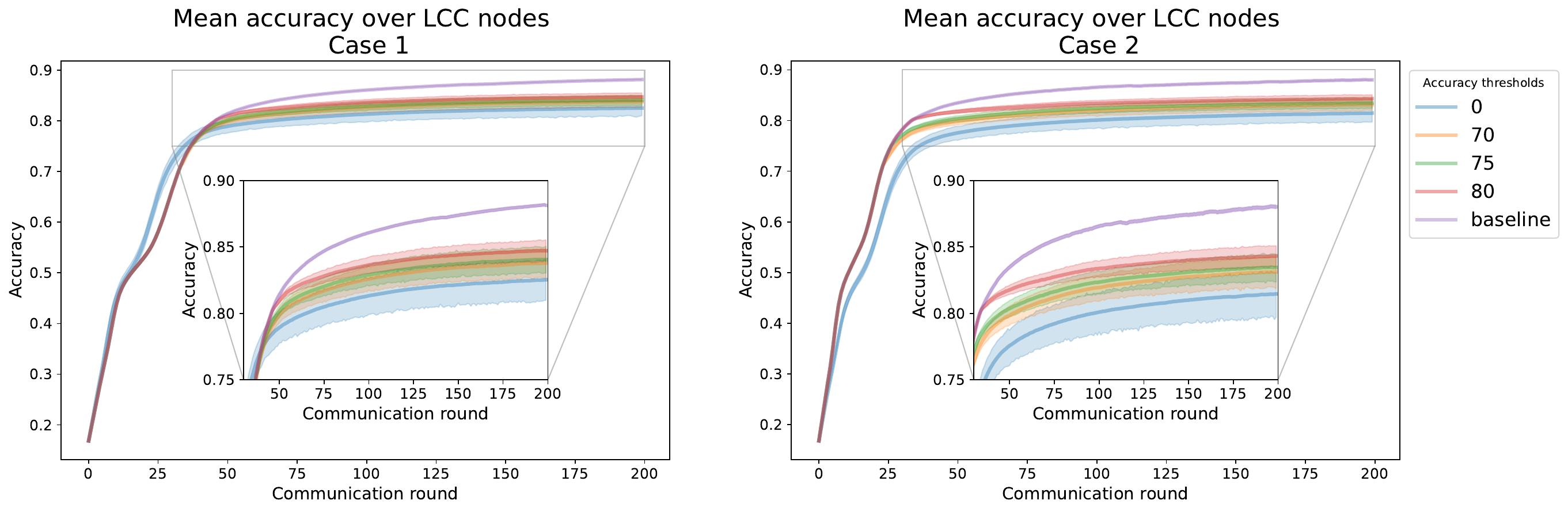}
    \caption{}
    \label{fig:c1c2_persistence_mean_acc_largecluster}
    \end{subfigure}
    \begin{subfigure}{\textwidth}
    \vspace{1em}
    \centering
    \includegraphics[width=\textwidth]{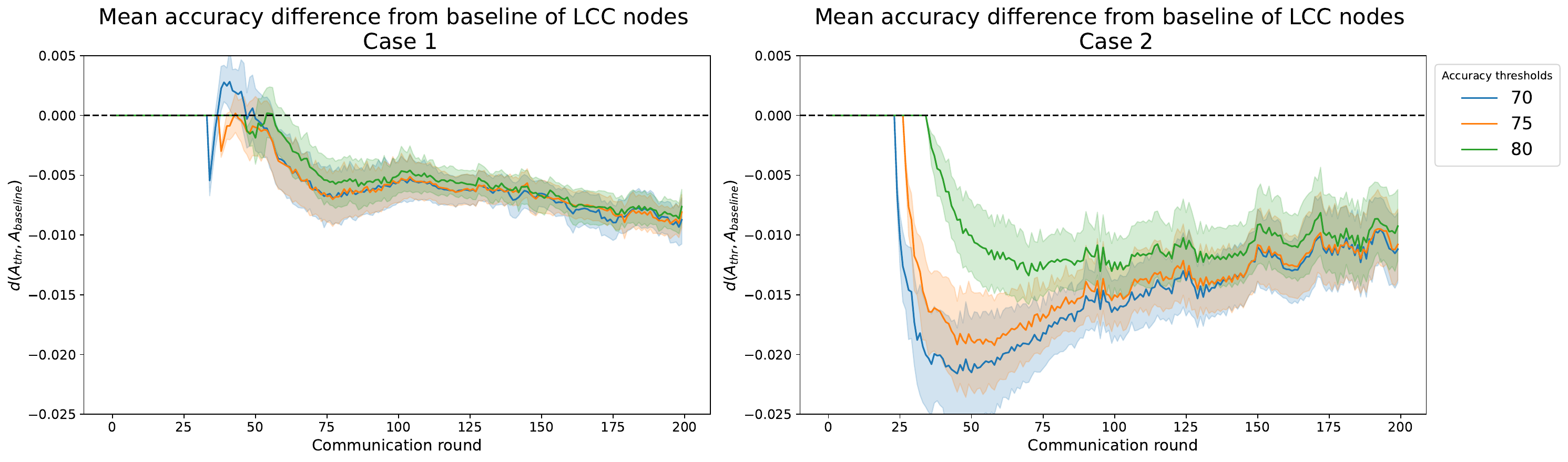}
    \caption{}
    \label{fig:c1c2_persistence_distance_baseline_largecluster}
    \end{subfigure}
    
    \caption{Knowledge persistence analysis (largest connected component). (a) Mean accuracy over all nodes in the largest connected component. (b) Mean accuracy difference with respect to baseline over all nodes in the largest connected component.}
    \label{fig:c1c2_persistence_largecluster} 
\end{figure}

First, we analyse the overall performance of the system, considering all nodes altogether (Figure~\ref{fig:c1c2_persistence_all}). To this end, in \Cref{fig:c1c2_persistence_mean_acc_all}  we show the average accuracy of the entire system, as defined in \Cref{def: mean_acc_definition}. As we can see from~\Cref{fig:c1c2_persistence_mean_acc_all}, there is a noticeable gap in accuracy with respect to the baseline. Furthermore, from the zoom inset, we can observe that, despite the curves being closely grouped together, there is a gap between the system's performances for the different accuracy thresholds. This can be better analysed by looking at the percentage difference (\Cref{fig:c1c2_persistence_distance_baseline_all}). With increasing accuracy threshold, the distance from the baseline decreases, and overall, it ranges anyway around 4 and 8\%. Also, in this case, we notice the slightly lower performance of Case 2 with respect to Case 1. Next, we focus individually on the two boundary cases of completely isolated nodes (Figure~\ref{fig:c1c2_persistence_isolated}), and on the case of the biggest connected component (Figure~\ref{fig:c1c2_persistence_largecluster}).

We start with isolated nodes. In \autoref{fig:c1c2_persistence_mean_acc_isolated}, we show the mean accuracy of the isolated nodes for different accuracy thresholds. We can observe several interesting features. First, we confirm the drop in accuracy after disruption suffered at any accuracy threshold (higher than 0). In~\ref{app:overfitting_isolated}, we discuss additional experiments that show that this drop is due to overfitting on the local data after disruption. Regardless of overfitting, this means that, for isolated nodes, it is fundamental to get access to knowledge extracted from data residing on other nodes and that they cannot fully compensate for the effect of a disruption using local data only. 
On the other hand, it is also interesting to observe that isolated nodes, at any accuracy threshold, achieve \emph{higher} accuracy with respect to the case where they started in isolation (accuracy threshold equal to 0). This means that, despite the observed drop, \emph{some} knowledge acquired thanks to collaborative learning still survives after disruption and persists for a long time. In other words, this knowledge \emph{persists} even in the absence of a large representation of labels in data residing locally. The difference metric (\autoref{fig:c1c2_persistence_distance_baseline_isolated} shows that the drop in accuracy ranges between 10 and 20\% depending on the accuracy threshold. All in all, \autoref{fig:c1c2_persistence_isolated} shows a significant persistence of knowledge after disruption even at isolated nodes, thanks to the fact that they had the chance of being exposed to the collaborative learning process before disruption, even though some loss of knowledge is unavoidable. The same behaviour can also be observed for small connected components and vanishes only when connected components contain an amount of data sufficient to recover from the loss of global knowledge diffusion implied from the time of disruption.

\autoref{fig:c1c2_persistence_largecluster} illustrates the average accuracy attained by nodes within the largest connected component and the percentage difference compared to the baseline. Intuitively, these nodes are less affected by disruptions, which is supported by the results. Specifically, we observe the following main findings. Firstly, large connected components demonstrate resilience to disruptions regardless of the accuracy thresholds, as evidenced by the nearly overlapping accuracy curves. This indicates that the timing of the disruption minimally impacts the performance of connected components. Secondly, there is a slight difference in performance compared to the baseline (approximately 1\%), suggesting that large connected components can effectively mitigate the effects of disruptions quite efficiently.
One particular aspect to note in this case is that the accuracy in Case 1 is maximised when the disruption occurs at time 0, and this effect is even more evident for earlier communication rounds. In other words, in Case 1, nodes of the largest connected component do not benefit at all from the presence of the other nodes in the network, which are actually detrimental to them. Albeit counter-intuitive, this behaviour can be explained also based on the analysis we have carried out in~\cite{palmieri2024impact}. Remember that in Case 1 the most central nodes (which undergo disruption) do not have data of their own, but they just average and forward models received from neighbours without any local retraining. Particularly early on in time, these nodes receive very heterogeneous models (as they are central, they typically have many neighbours, see Section~\ref{sec:settings_disruption_analysis}) and therefore, the average model they compute is not very accurate unless it has a chance of being retrained on local data. As a side effect, when disruption occurs after time 0, central nodes inject in the largest (after disruption) connected component ``noisy" models, which do not contribute to, yet work against, the overall accuracy achieved by nodes in this connected component.

\emph{Take-home:} Considering the results presented in this section altogether, we can observe a significant persistence of knowledge after disruption in all cases. Even in the most unfavourable conditions (i.e., isolated nodes) the accuracy achieved sufficiently long after the disruption is significantly higher than what would be achieved if those nodes were never exposed to the decentralised learning process (i.e., the case ``accuracy threshold 0"). There is clearly a loss of accuracy with respect to the case where no disruption occurs, but this is rather limited (in the order of 20\% maximum in our experiments). Moreover, if nodes remain grouped in sufficiently large connected components, the decentralised learning process is extremely robust, and, if sufficient time is allowed after disruption, the achieved accuracy is basically indistinguishable from the case when no disruption occurred.

\subsection{The effect of non-IID data}
\label{sec:results_case3}

In the scenarios considered so far, nodes have (as summarized in~\autoref{tab:summary_simulations}) either the same amount of images per class or none at all (disrupted nodes in Case 1), so the data distribution was IID assuming that data were present. This implied that each node with data would contribute equally to the learning process. In this final set of results, we want to focus, instead, on the effects of an unequal data distribution. As we explained in Section~\ref{sec:settings_scenarios}, in Case 3, data labels are divided into two classes, $\mathcal{L}_1$ and $\mathcal{L}_2$, and we considered two different sub-scenarios regarding their distribution. In both sub-scenarios, the nodes will be divided into two groups, G1 and G2, based on the amount of images belonging to $\mathcal{L}_2$ they are assigned. $\mathcal{L}_1$ images are distributed in an IID fashion across all nodes by simply splitting the approximately 6,000 images among all nodes (hence, each node gets around 60 images per class). Instead, $\mathcal{L}_2$ images are distributed as follows: only 10, 20 or 30 images belonging to $\mathcal{L}_2$ are assigned to the G1 nodes, while all the others are split equally among G2 nodes. In the first sub-scenario, G2 nodes coincide with the disrupted nodes, which end up with approximately 500 images per $\mathcal{L}_2$-class. Thus, in Case 3.1, disrupted nodes are disproportionately better at classifying $\mathcal{L}_2$-images and losing them after the disruption is expected to significantly impact the learning process.
In the second sub-scenario, G2 nodes will be the 10 nodes with the lowest structural hole score, meaning that, after the disruption, among the surviving nodes there will be ones that preserve more knowledge on the $\mathcal{L}_2$-classes. 
Before we continue, note that in Case 3 we had to consider higher accuracy thresholds than in the previous cases (87\%, 90\%, 92\%). This is because in this Case 3 configuration, there are much more data around in the network overall, so accuracies of 80\% are basically reached in the very first communication rounds. 

\subsubsection{Case 3.1: Disrupted nodes with more knowledge}
\label{sec:results_3_1}

\begin{figure}[ht!]
    \centering
    \includegraphics[width=\textwidth]{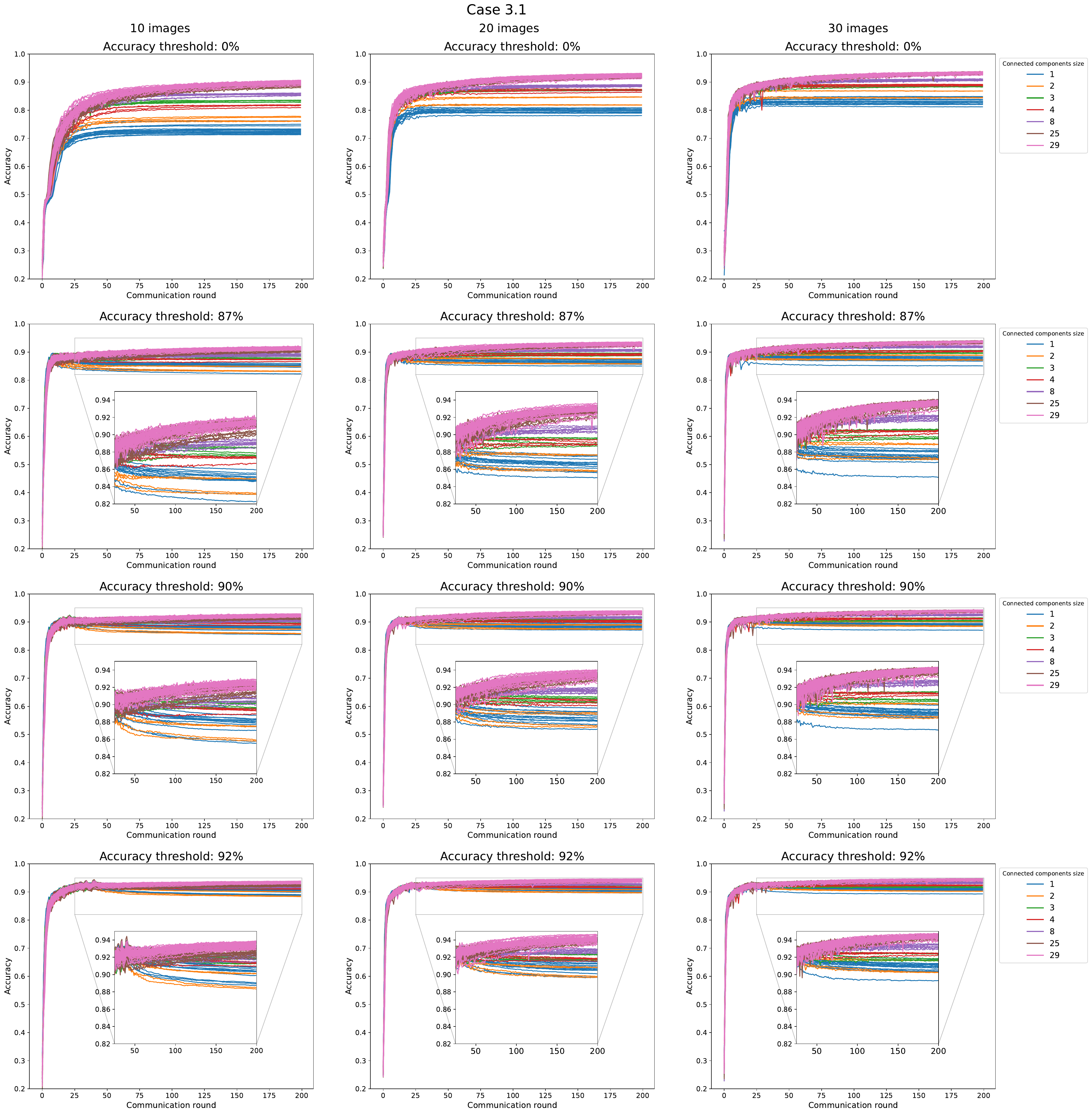}
    \caption{Accuracy curves for each active node in Case 3.1. Top to bottom:increasing accuracy threshold condition for the disruption. From left to right: increasing value of the number of images in the $\mathcal{L}_2$-label.
    The curves are coloured based on the size of the connected component to which the corresponding node belongs. The inset zoom displays the clusterization of accuracy levels.}
    \label{fig:c3_accuracy_over_time}
\end{figure}

Figure~\ref{fig:c3_accuracy_over_time} displays the accuracy over time for Case 3.1. Let us begin with the most challenging configuration (first column), where the surviving nodes after the disruption only possess 10 images for each $\mathcal{L}_2$-label (those ranging from 5 to 9 in the MNIST dataset). When the disruption occurs at time zero, the spread of curves is quite broad, indicating that some nodes lack the ability to learn solely based on local data. As the disruption is delayed, the spread narrows, indicating that even the most isolated nodes post-disruption can benefit from collaborative training that occurred before disruption, despite having very few local $\mathcal{L}_2$-images. Note, in the inset of the figure, that more peripheral nodes experience a drop in accuracy after the disruption, as observed in previous cases. However, the accuracy remains significantly higher than when there is no collaboration at all (corresponding to an accuracy threshold of 0\%). With an increase in the number of local images (second and third columns), we observe an overall facilitation in the learning process, as expected. In this scenario, the timing of the disruption plays a less significant role, as the local data suffice to compensate for the loss of collaborators.

\begin{figure}[t!]
    \centering
    \includegraphics[width=\textwidth]{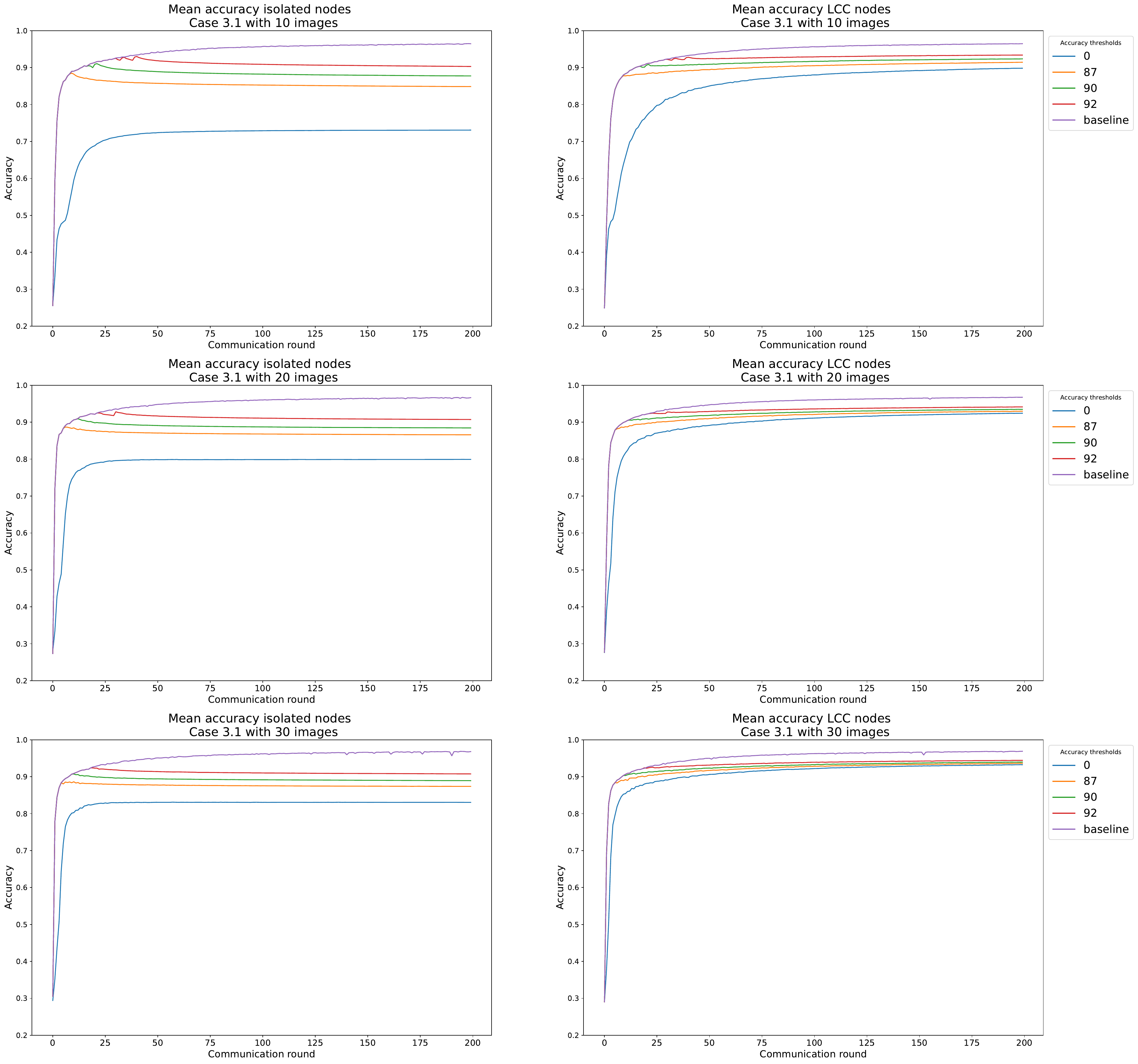}
    \caption{Mean accuracy in Case 3.1 over isolated nodes (left) and over nodes belonging to the largest connected component (right), for different disruption thresholds and number of $\mathcal{L}_2$-images.}
    \label{fig:c3_isolated_and_large}
\end{figure}

\begin{figure}[t!]
    \begin{subfigure}{\textwidth}
        \includegraphics[width=\textwidth]{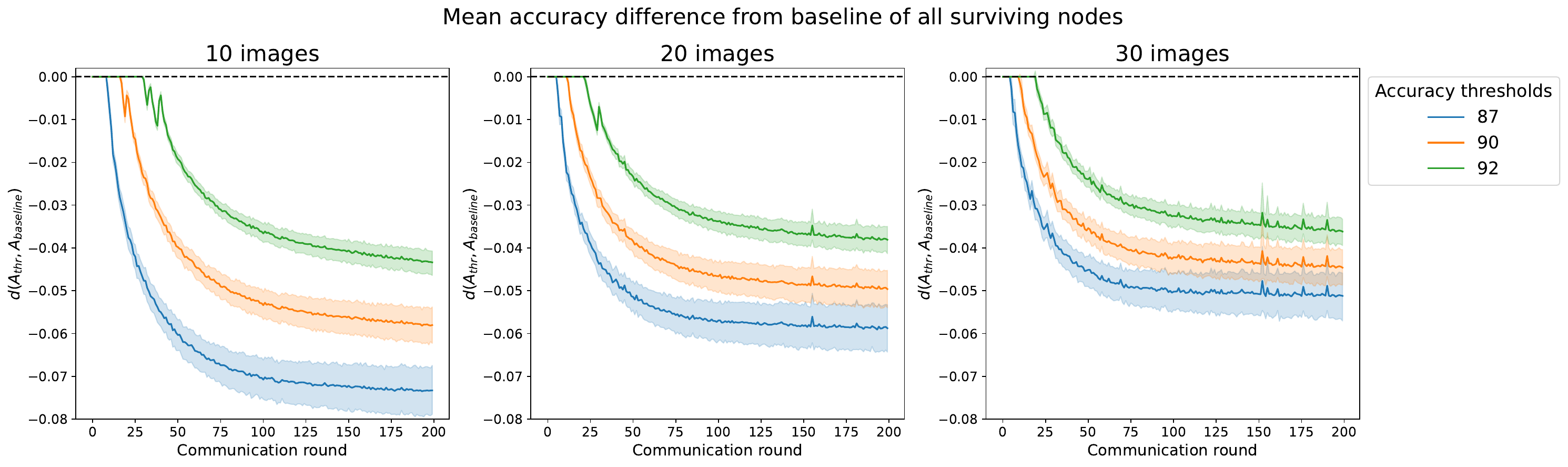}
    \end{subfigure}
    \begin{subfigure}{\textwidth}
        \includegraphics[width=\textwidth]{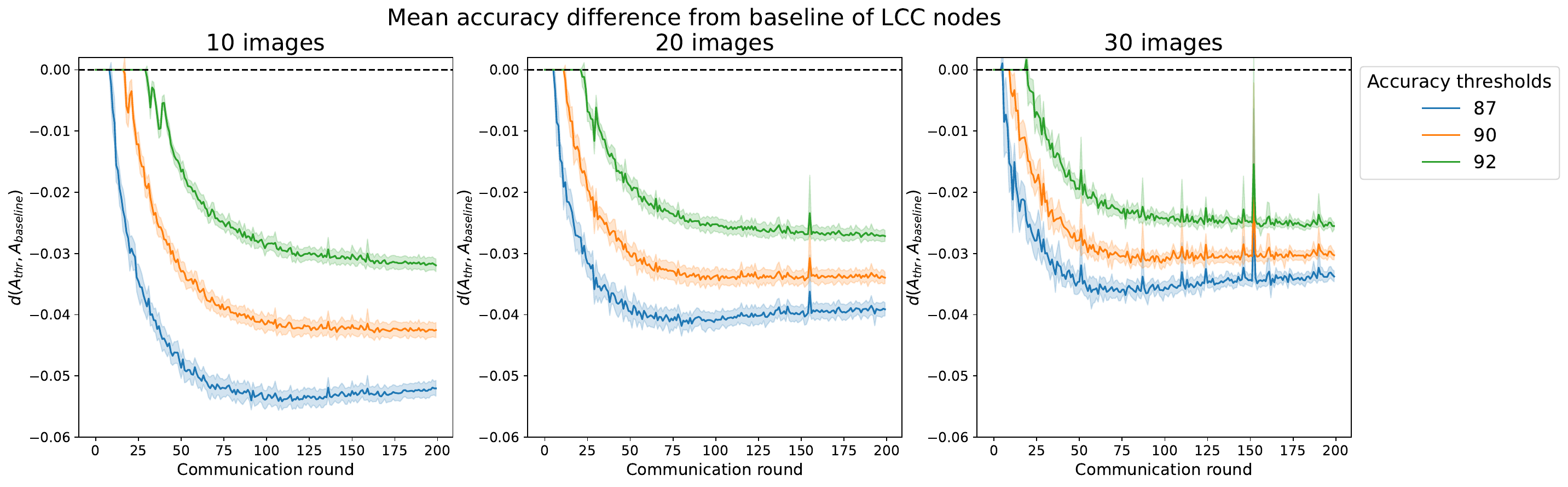}
    \end{subfigure}
    \begin{subfigure}{\textwidth}
        \includegraphics[width=\textwidth]{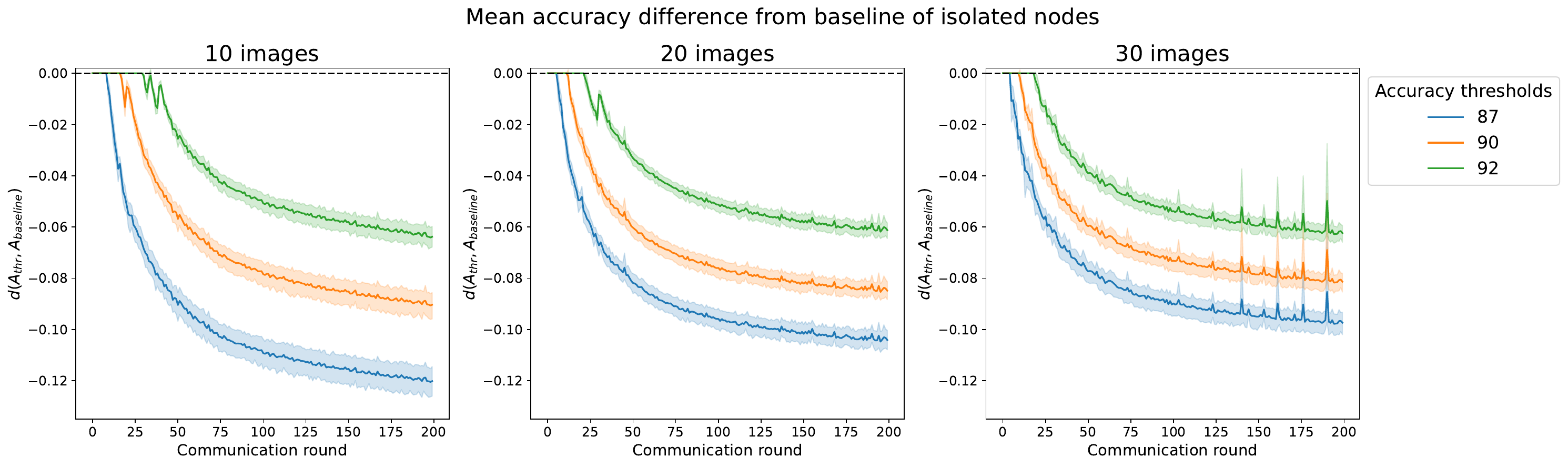}
    \end{subfigure}
    \caption{Accuracy difference in Case 3.1 with respect to baseline for: all surviving nodes (top row), largest connected component (middle row), isolated nodes (bottom row), for different disruption thresholds.}
    \label{fig:c3_difference_baseline}
\end{figure}

Zooming in on the differences between isolated nodes and the largest connected component after the disruption (Figure~\ref{fig:c3_isolated_and_large}), we confirm that nodes that become completely disconnected post-disruption can capitalise on the knowledge received and assimilated before the disruption. Collaborative learning benefits them most when they have very few local training images initially. The largest connected component (second column of Figure~\ref{fig:c3_isolated_and_large}) suffers much less from the disruption because it can effectively pool its nodes' knowledge thanks to the remaining graph connectivity. However, similar to previous cases, later disruptions and an increase in local images also alleviate the burden on their learning process. These findings are clearly confirmed when looking at the differences in accuracy (\autoref{fig:c3_difference_baseline}).

\emph{Take-home:} The results for Case 3.1 confirm the robustness of the decentralised learning process to network disruptions: as long as some knowledge is available and accessible through the communication graph, compensating for the loss of collaborators -- even excellent ones like in Case 3.1 --  is feasible and the loss in accuracy with respect to the baseline of no disruption is drastically mitigated. More than other, nodes that lose \emph{all} collaborators benefit significantly from the accumulation of deep knowledge circulated before the disruption.

\subsubsection{Case 3.2: Peripheral nodes with more knowledge}
\label{sec:results_3_2}

\begin{figure}[p]
    \centering
    \includegraphics[width=\textwidth]{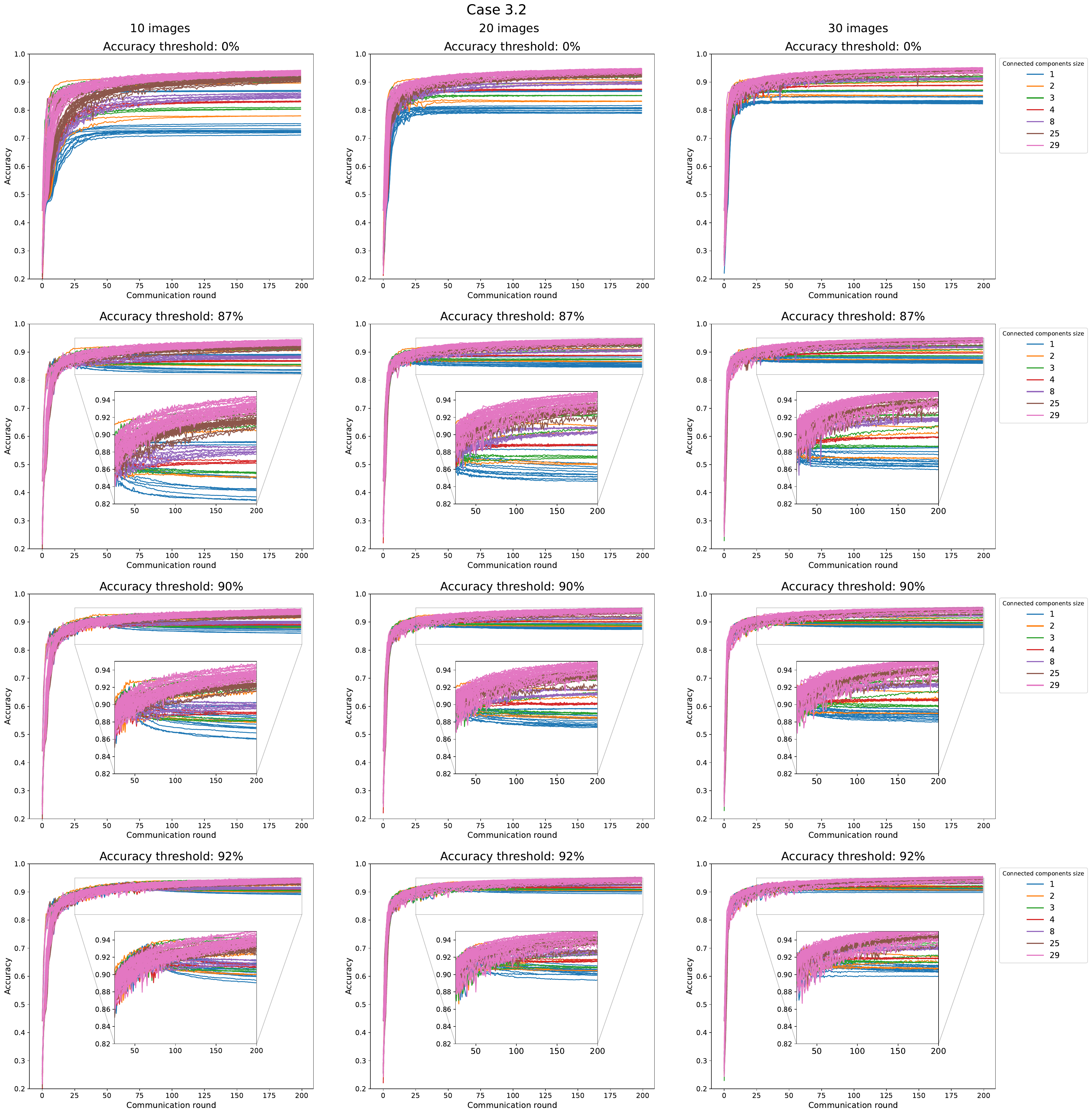}
    \caption{Accuracy curves for each active node in Case 3.2. Top to bottom: increasing accuracy threshold condition for the disruption. From left to right: increasing value of the number of images in the $\mathcal{L}_2$-label. The curves are coloured based on the size of the connected component to which the corresponding node belongs. The inset zoom displays the clusterization of accuracy levels.}
    \label{fig:c3_peripheral_accuracy_over_time}
\end{figure}

\begin{figure}[ht]
    \centering
    \includegraphics[width=\linewidth]{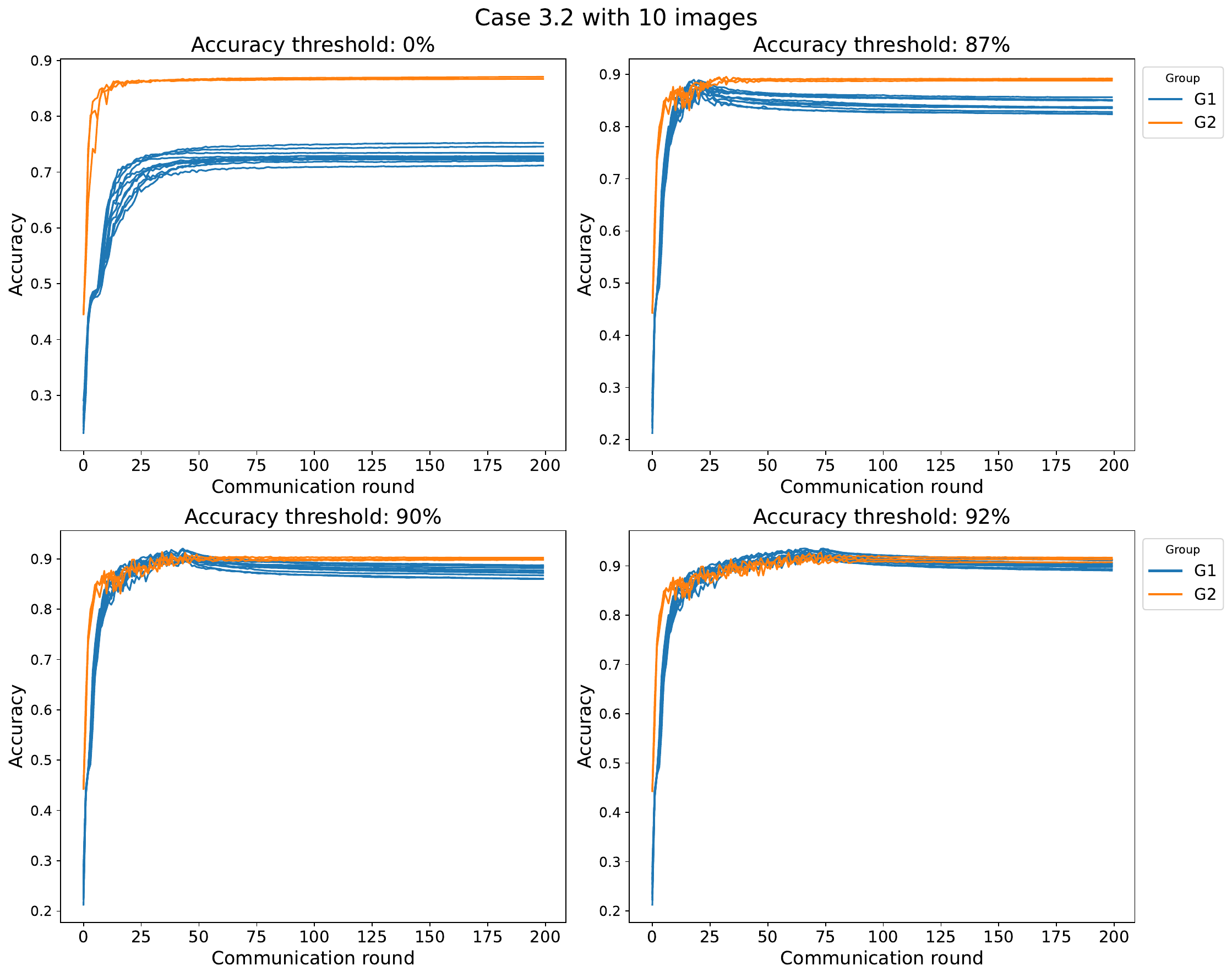}
    \caption{Accuracy over time in Case 3.2 for the isolated nodes. Coloring is based on their group membership. From left to right: increasing accuracy threshold. Top to bottom: increasing value of the number of images in the $\mathcal{L}_2$-label.}
    \label{fig:c3_peripheral_special_nodes}
\end{figure}

\autoref{fig:c3_peripheral_accuracy_over_time} shows the accuracy over time for Case 3.2, which follows the same behavior as that of Case 3.1, with the narrowing of the curves as the disruption is delayed proving the impact of collaborative learning. A notable distinction arises in Case 3.2, where some of the blue curves — which represent isolated nodes — exhibit higher accuracy levels compared to others. This phenomenon can be attributed to the fact that some of the isolated nodes belong to the G2 group (with a high number of samples for $\mathcal{L}_2$-data). \autoref{fig:c3_peripheral_special_nodes} shows the accuracy progression over time for isolated nodes when G1 nodes have access to 10 images of the $\mathcal{L}_2$ class, with G2 nodes distinctly highlighted. Moreover, going from left to right, we can observe how the distance in performance of the G1 and G2 isolated nodes evolves w.r.t the delay in the disruption. We can see that the G1 nodes eventually reach comparable accuracy level of the G2 nodes. This convergence in performance clearly indicates that delaying the disruption of collaborative learning benefits the G1 isolated nodes, allowing these nodes to compensate for their smaller amount of $\mathcal{L}_2$ data. This is the effect of knowledge persistence. The extended interaction period helps G1 nodes to leverage the collaborative learning, improving their ability to perform well even after their isolation. 

\subsubsection{Comparison between Case 3.1 and Case 3.2}
\label{sec:results_31_32}

\begin{figure}[ht!]
    \centering
    \includegraphics[width=\linewidth]{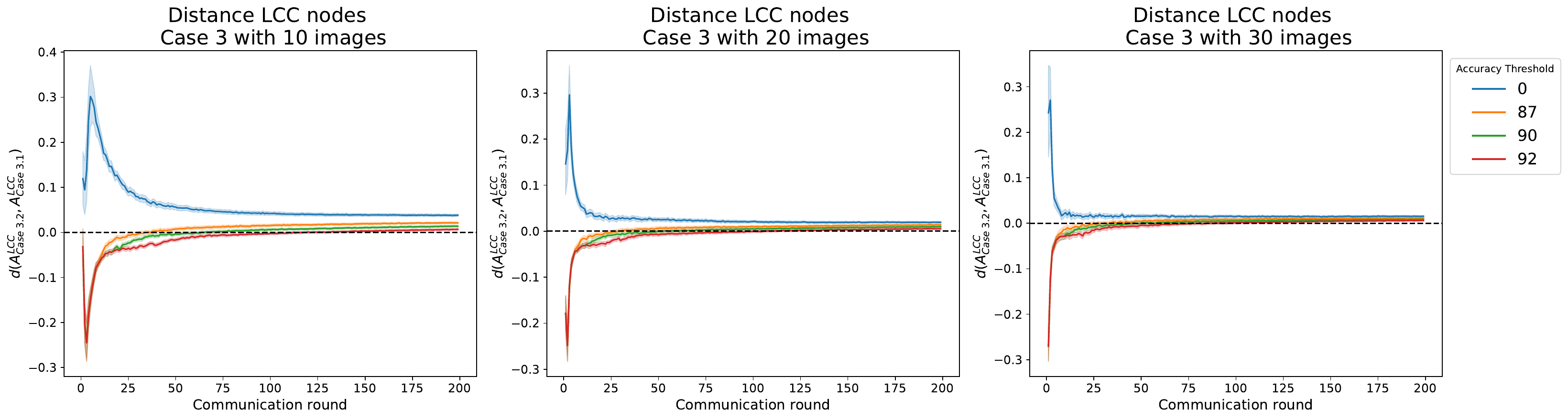}
    \caption{Accuracy distance of nodes in the largest (after disruption) connected component between Case 3.1 and Case 3.2. A positive difference implies that the accuracy of Case 3.2 is higher than that of Case 3.1, and vice versa. The dotted line represents the equivalence between the two cases.}
    \label{fig:c3_comparison}
\end{figure}

The major difference between the two sub-scenarios is that some G2 nodes will survive the disruption in Case 3.2. This difference influences the behaviour of the system, as seen when comparing the accuracy levels of nodes within the largest connected component (LCC) between Case 3.1 and Case 3.2, as shown in \autoref{fig:c3_comparison}. To this purpose, we use Definition~\ref{def:accuracy_diff}. Initially, while the network remains fully connected, Case 3.1 exhibits higher performance than Case 3.2, reflected in a negative distance between the accuracies. This suggests that during the collaborative phase, nodes in Case 3.1 are able to benefit more from the shared learning process, because more knowledge is placed on the nodes that are more central, hence the spreading of knowledge is faster. However, as time progresses and the network experiences disruption, the trend shifts. In the long term, nodes in Case 3.2 achieve higher accuracy levels, mainly due to the sustained presence of G2 nodes, which ensures a greater overall availability of data within the network.
Moreover, as the number of $\mathcal{L}_2$ images accessible locally to G1 nodes increases (from left to right), the gap in accuracy between the two cases narrows (i.e., the difference approaches zero). This indicates that the collaborative learning process within the LCC helps to mitigate the initial disadvantage faced by G1 nodes due to their smaller data sets. Essentially, the prolonged interaction with other nodes allows G1 nodes to better utilize the data available through collaboration, partially compensating for the lack of direct access to $\mathcal{L}_2$-data. This is especially notable by looking at the blue curve, where the system starts as already disrupted in both cases. 

\emph{Take-home:} In the long run, after a disruption, the learning process is more effective when more data are retained within the surviving network, even if they reside on non-central nodes. Consequently, in Case 3.2, the final accuracy is higher than in Case 3.1. However, the difference is not large and decreases the later the disruption occurs and the larger the local datasets become. Additionally, the same amount of data stored in central nodes rather than in peripheral nodes accelerates the learning process, as demonstrated by the superior initial (pre-disruption) performance of Case 3.1 compared to Case 3.2.

\section{Conclusions and future work}
\label{sec:conclusions}

In this paper we have focused on fully decentralised learning, and we have analysed the robustness of the learning process in case of disruptions. Specifically, after a variable initial period where the network is intact and, therefore, the decentralised process can proceed without disruptions, we have removed a given percentage of central nodes, and we have analysed, over time, the accuracy achieved by the surviving nodes. Specifically, we have considered various types of disruptions involving loss of either connectivity alone or connectivity and data together.

The most important finding we can highlight is that decentralised learning appears to be remarkably robust to all types of disruptions considered, as long as surviving nodes hold a sufficient fraction of representative data to sustain the learning process after disruption. For nodes belonging to large connected components (after disruption) the loss of accuracy is negligible compared to the case when no disruption occurred. For isolated nodes, the loss of accuracy is larger, but we have never observed it exceeding 20\% with respect to the case of no disruption. In all cases, we have shown that if surviving nodes are part of the decentralised learning process for some time before disruption occurs, then they can retain most of the knowledge acquired before disruption and achieve a much higher accuracy than in the case when the network starts already in the same configuration as after disruption.

We identify three key reasons for this behaviour. First, knowledge acquired before disruption persists, and is not lost even by isolated nodes, as long as they have even a small local dataset to refresh it through local training. Second, accuracy can be recovered if data is present ``somewhere" in the network, even though very distributed across surviving nodes. Third, even modest connectivity supports efficient recovery from failures, as nodes in connected components, thanks to the collaborative nature of decentralised learning, are able to jointly train very accurate models even after the disruption.

This work, along with~\cite{palmieri2023exploring}, marks an initial step toward a deeper understanding of the interplay between the topology and dynamics of complex systems and the learning processes they support. However, many important research questions remain unexplored.
A natural extension of this work would be to explore more sophisticated learning strategies beyond DecAvg, and to apply the framework to more challenging datasets than the baseline MNIST. Additionally, the network topologies studied could be expanded to include more real-world graphs, as well as synthetic ones that are less representative of real systems (e.g., regular graphs). Analyzing these diverse topologies could help establish upper and lower bounds on robustness as function of different network classes.
From the disruption standpoint, this work focused on the failure of entire nodes at a specific moment in time. However, future research could explore more dynamic failure models in which failures occur gradually over time, potentially affecting edges rather than nodes, or causing nodes to fail sequentially instead of abruptly all at once. We also assume that nodes do not reconnect once disrupted, whereas in real-world scenarios, disruptions might be temporary. A deeper analysis could investigate the impact of temporary disruptions, especially in heterogeneous computing environments where nodes operate at varying speeds, since the disruption of faster versus slower nodes could affect the learning process differently.
Another important direction is the theoretical derivation of robustness properties, rather than solely inferring them from experiments, which could offer a more formal understanding of the system’s behavior. 
Additionally, an orthogonal yet critical research avenue is robustness against malicious threats, rather than just failures, a topic that remains largely unexplored in the current literature on fully decentralised learning.

\section*{Acknowledgments}
\noindent
This work was partially supported by the H2020 HumaneAI Net (952026) and by the CHIST-ERA-19-XAI010 SAI projects. C. Boldrini's and M. Conti's work was partly funded by the PNRR - M4C2 - Investimento 1.3, Partenariato Esteso PE00000013 - ``FAIR'', A. Passarella's and L. Valerio's work was partially supported by the European Union - Next Generation EU under the Italian National Recovery and Resilience Plan (NRRP), Mission 4, Component 2, Investment 1.3, CUP B53C22003970001, partnership on “Telecommunications of the Future” (PE00000001 - program “RESTART”). J. Kert\'esz acknowledges support also from ERC grant No. 810115-DYNASNET.

\bibliographystyle{elsarticle-num} 
\bibliography{references}

\appendix
\section{Overfitting on isolated nodes}
\label{app:overfitting_isolated}
\begin{figure}[h]
    \centering
    \includegraphics[width=1.2\textwidth]{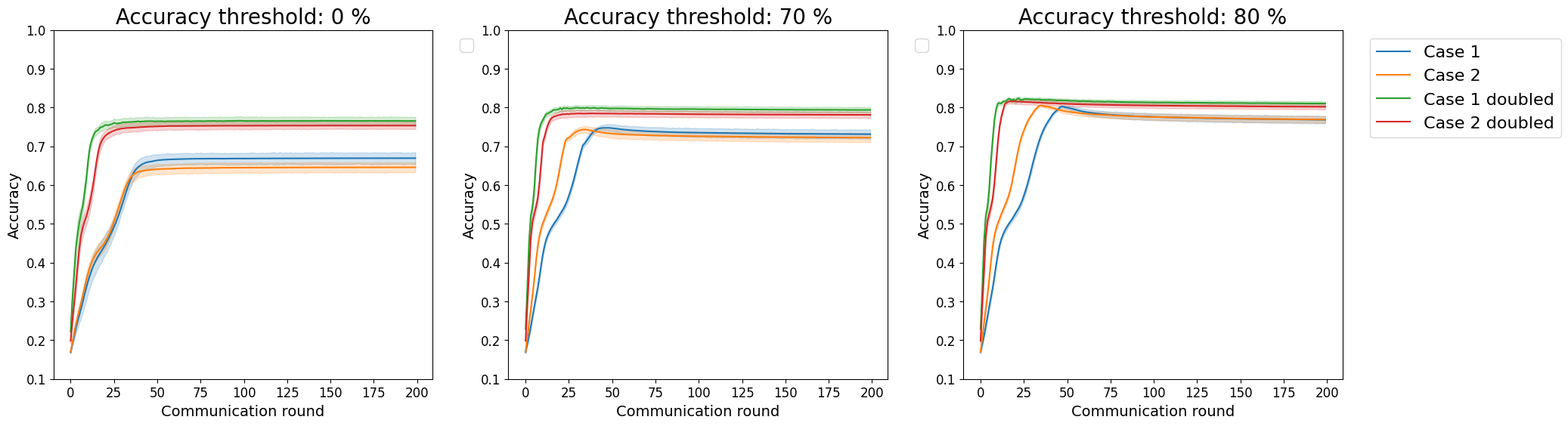}
    \caption{Mean accuracy curves of the isolated nodes for Case 1 and Case 2 and counterparts with doubled local training set size.}
    \label{fig:case12_doubled_comparison}
\end{figure}

%
When looking at the accuracy curves of isolated nodes (Figure~{\ref{fig:c1c2_persistence_isolated}}), we often noticed a drop in accuracy after a disruption. This drop observed in isolated nodes can be traced back to the limited number of images per class available in the individual datasets. 
As indicated in Table~{\ref{tab:summary_simulations}}, the individual datasets contain either 7 images per class (Case 1) or 6 images per class (Case 2). This small number of images per class may lead to overfitting when nodes become completely isolated. With such limited data, after disruption, the local model begins to memorize the specific properties of the few available images rather than learning general patterns, which gradually reduces its ability to generalize effectively to new data (i.e., those in the test set). To prove this, we conducted an additional set of experiments where we doubled the number of images per class. The results, presented in Figure~{\ref{fig:case12_doubled_comparison}}, show the mean accuracy curves for all isolated nodes in both Case 1 and Case 2, comparing the small local training dataset used in the previous experiments with larger versions. It is evident that the significant drop in accuracy observed with the smaller datasets disappears when the local data availability is increased. While a decreasing trend in accuracy remains, the overfitting problem is notably mitigated, highlighting the importance of sufficient data for generalization in isolated nodes.

\section{Karate Club Graph}
\label{app:karate_club}

\begin{figure}[ht]
\centering
    \begin{subfigure}{.35\textwidth}
    \centering
    \includegraphics[width=\textwidth]{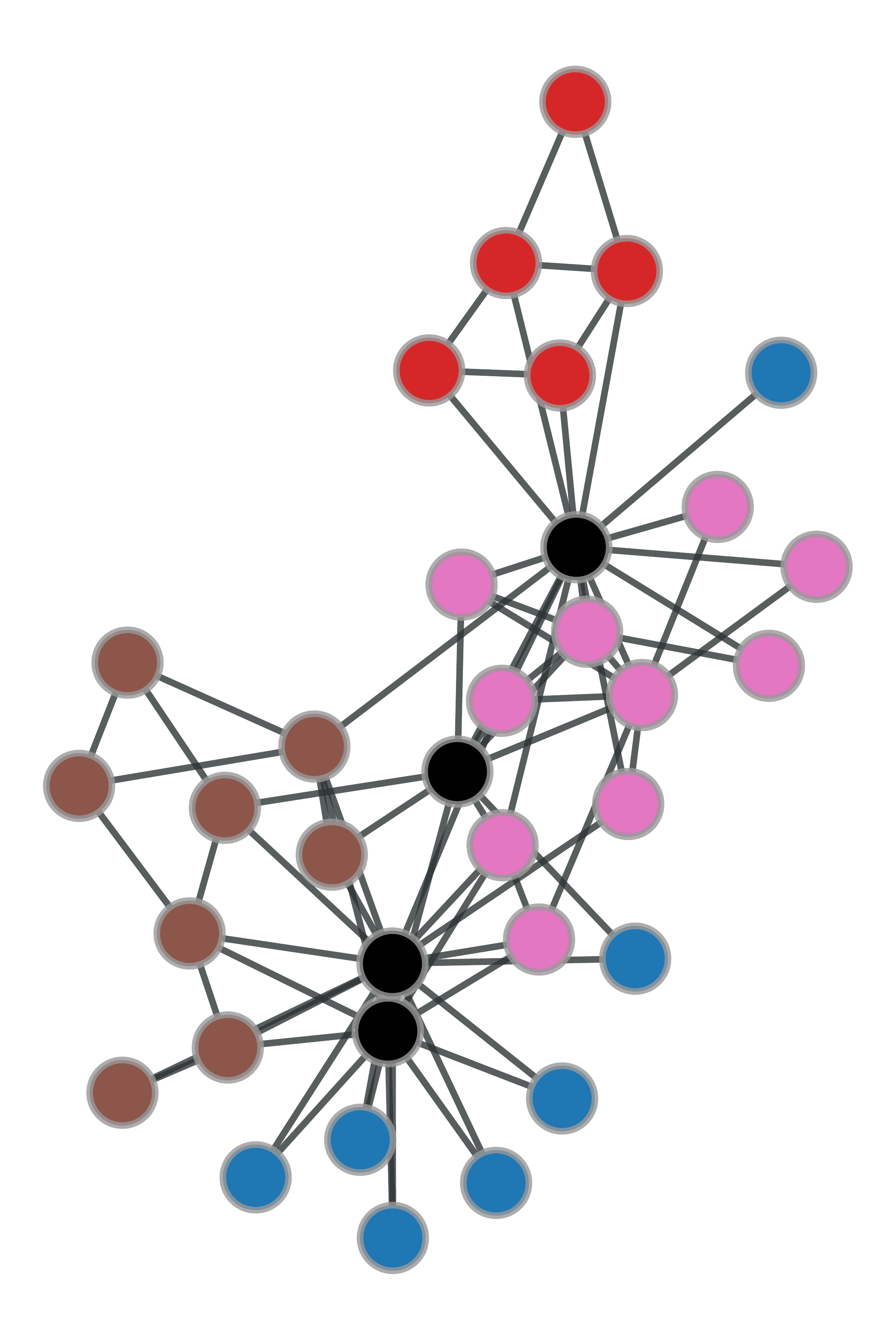}
    \caption{}
    \end{subfigure}%
    \begin{subfigure}{.35\textwidth}
    \centering
    \includegraphics[width=\textwidth]{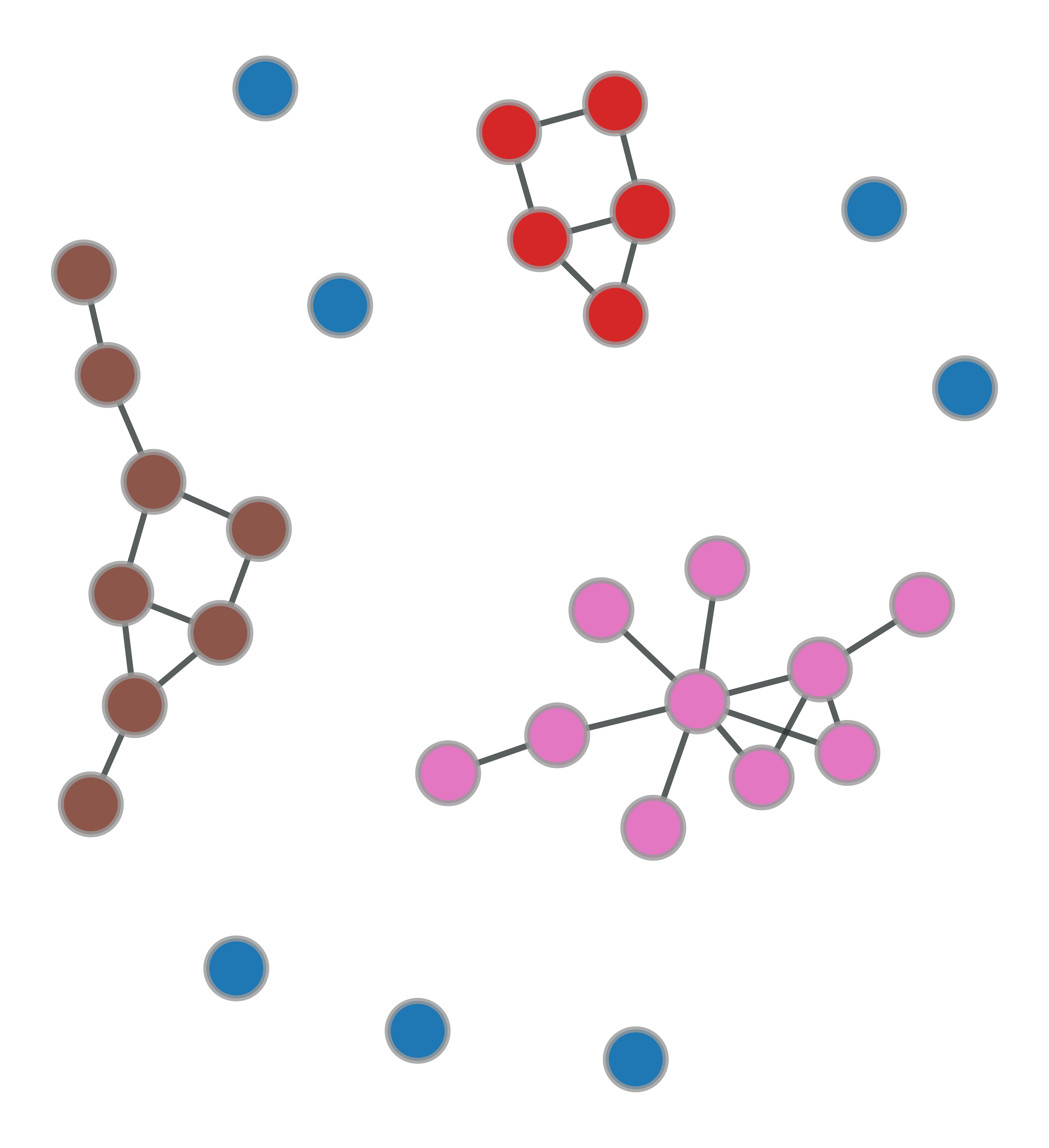}
    \caption{}
    \end{subfigure}
\caption{Karate club network before (left) and after (right) the disruption. The colouring of nodes is based on the size of the connected component they belong to after disruption, using as basis the connected components' coloring used for the Barabasi-Albert. The black nodes in the left image are the nodes that will be switched off when the disruption condition is fulfilled.}
\label{fig:ka_net_bef_aft}
\end{figure}

\begin{figure}[ht]
    \centering
    \includegraphics[width=0.5\textwidth]{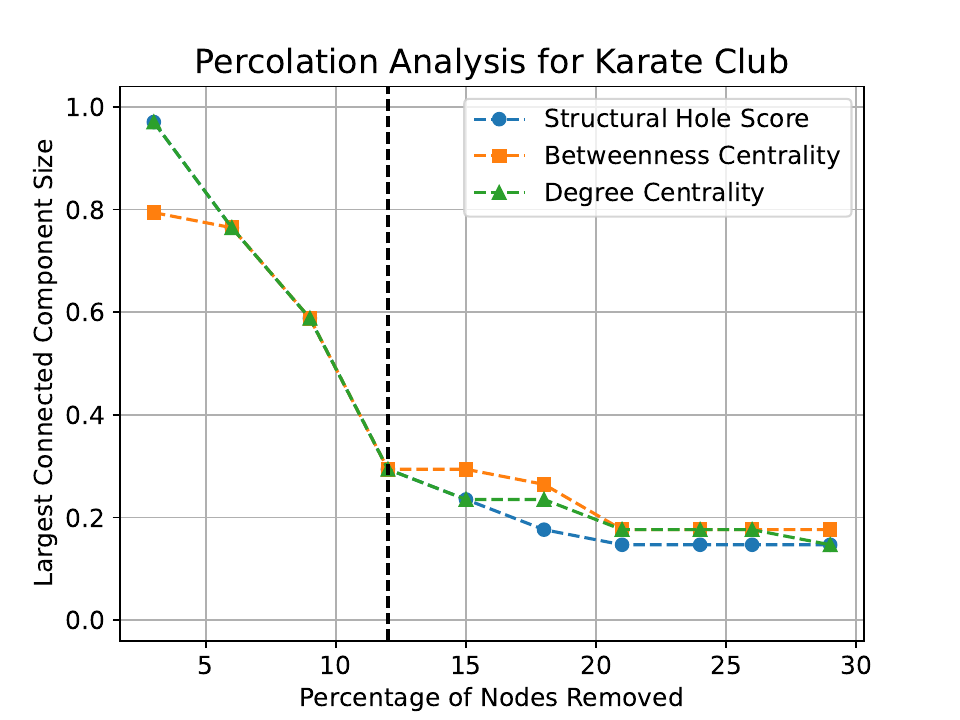}
    \caption{Karate Club Graph's size of the largest connected component as nodes are removed progressively. The dotted line represent the percentage chosen to be removed to account for the same LCC size of the BA network.}
    \label{fig:ka_perc}
\end{figure}

\begin{figure}[ht]
\begin{adjustwidth}{-1cm}{-1cm}
    \begin{subfigure}{1.2\textwidth}
        \includegraphics[width=\textwidth]{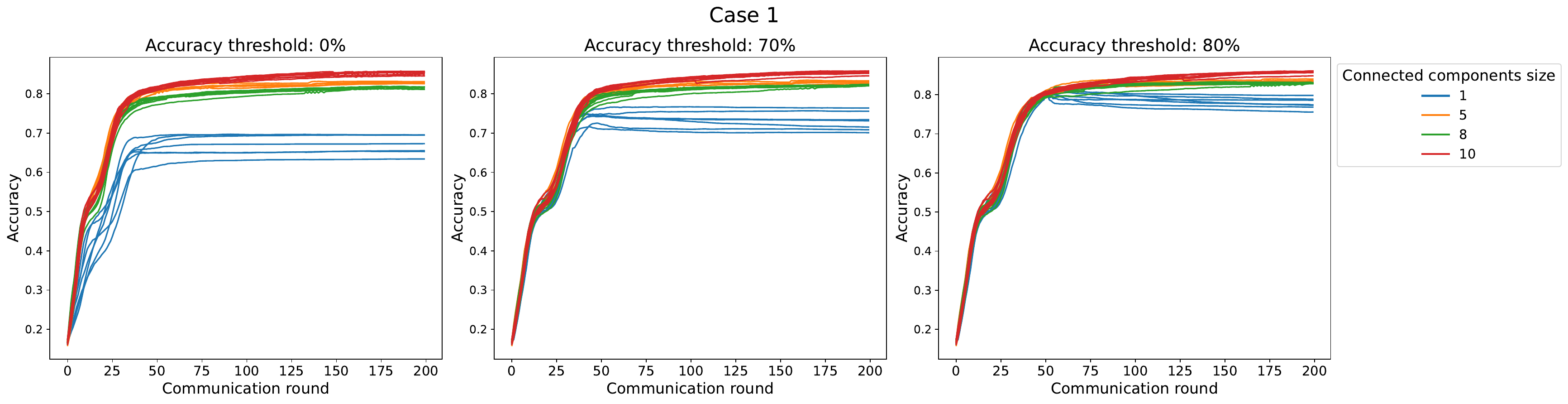}
        \caption{}
        \label{fig:ka_case1_accuracy}
    \end{subfigure}
    
    \begin{subfigure}{1.2\textwidth}
        \includegraphics[width=\textwidth]{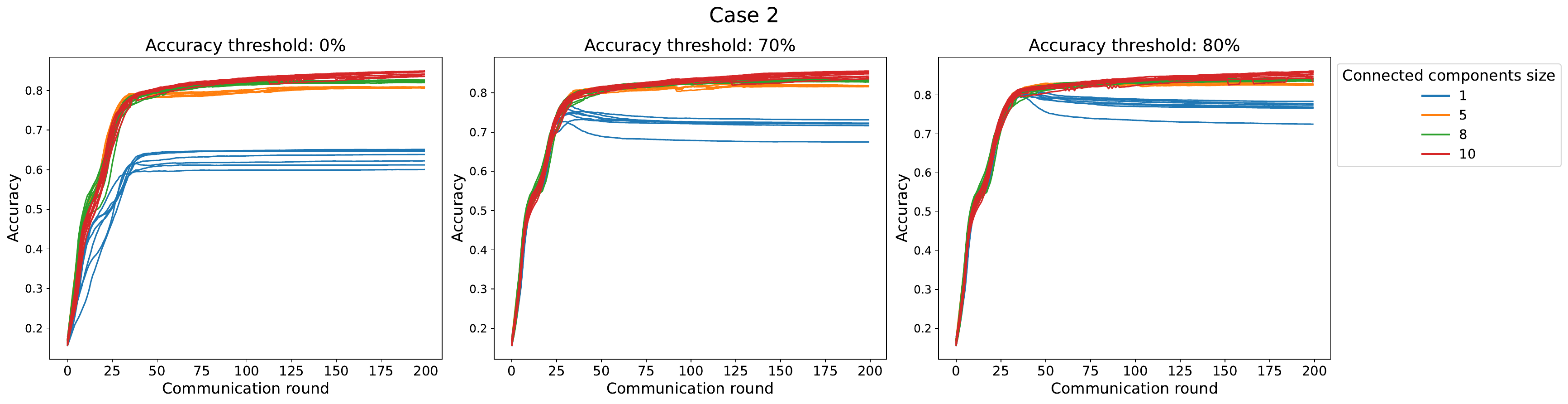}
        \caption{}
        \label{fig:ka_case2_accuracy}
    \end{subfigure}
    \end{adjustwidth}
    \caption{Accuracy curves for each surviving node in Case 1 (a) and Case 2 (b). From left to right: increasing accuracy threshold condition for the disruption. The curves are coloured based on the size of the connected component to which the corresponding node belongs after the disruption. Accuracy threshold 75\% is omitted for ease of visualisation, as it only showed an intermediate behaviour between threshold 70\% and 80\%.}
    \label{fig:ka_accuracy_all}
\end{figure}

\begin{figure*}[ht]
    \centering
    \begin{subfigure}{\textwidth}
    \centering
    \includegraphics[width=\textwidth]{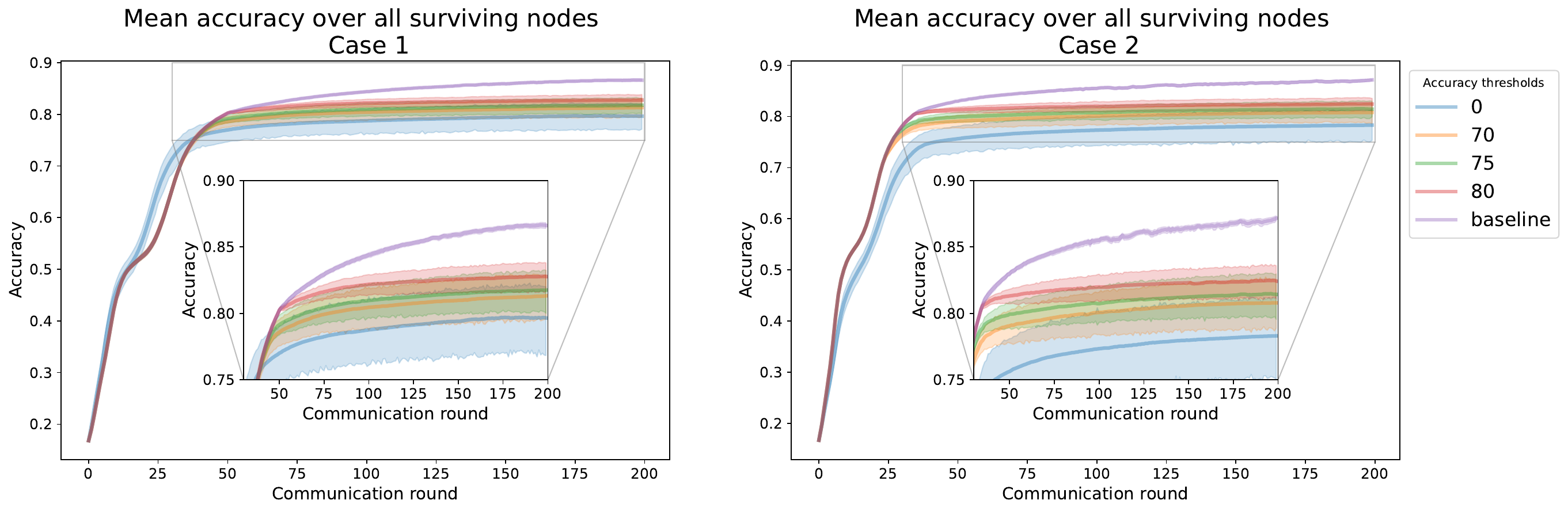}
    \caption{}
    \label{fig:ka_c1c2_persistence_mean_acc_all}
    \end{subfigure}
    \begin{subfigure}{\textwidth}
    \centering
    \includegraphics[width=\textwidth]{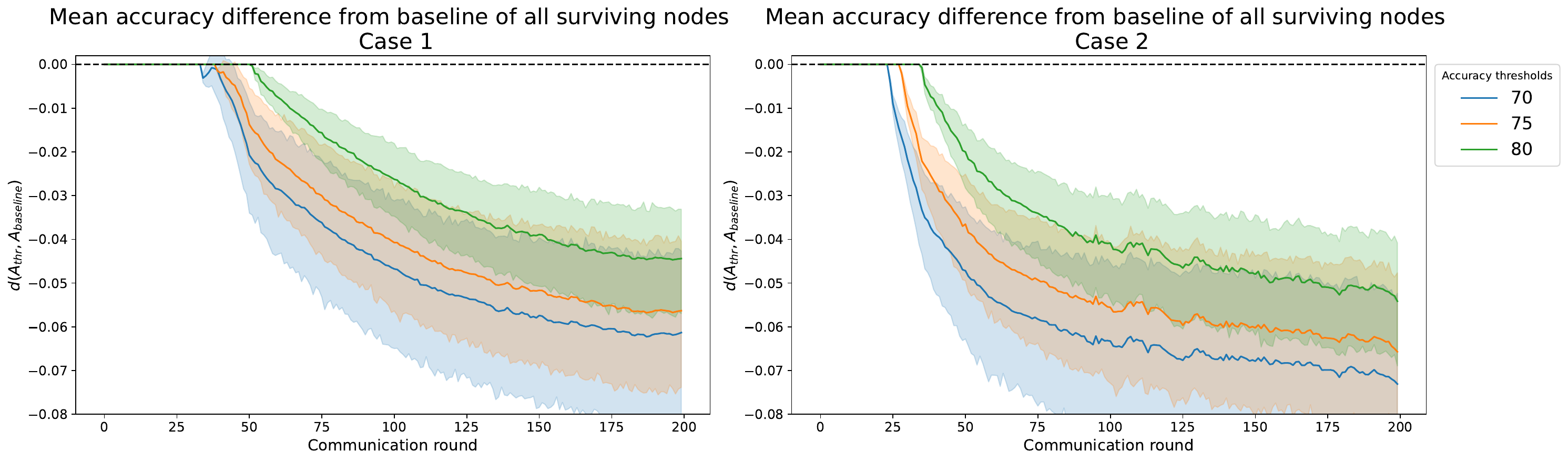}
    \caption{}
    \label{fig:ka_c1c2_persistence_distance_baseline_all}
    \end{subfigure}
    \caption{Knowledge persistence analysis (all nodes). (a) Average accuracy of the system calculated among all non-switched-off nodes for different accuracy thresholds. (b) Distance between the different accuracy thresholds and the baseline. The distance is calculated as the percentual difference between the accuracies.}
    \label{fig:ka_c1c2_persistence_all}
\end{figure*}

\begin{figure}[ht]
    \centering
    \begin{subfigure}{\textwidth}
    \centering
    \includegraphics[width=\textwidth]{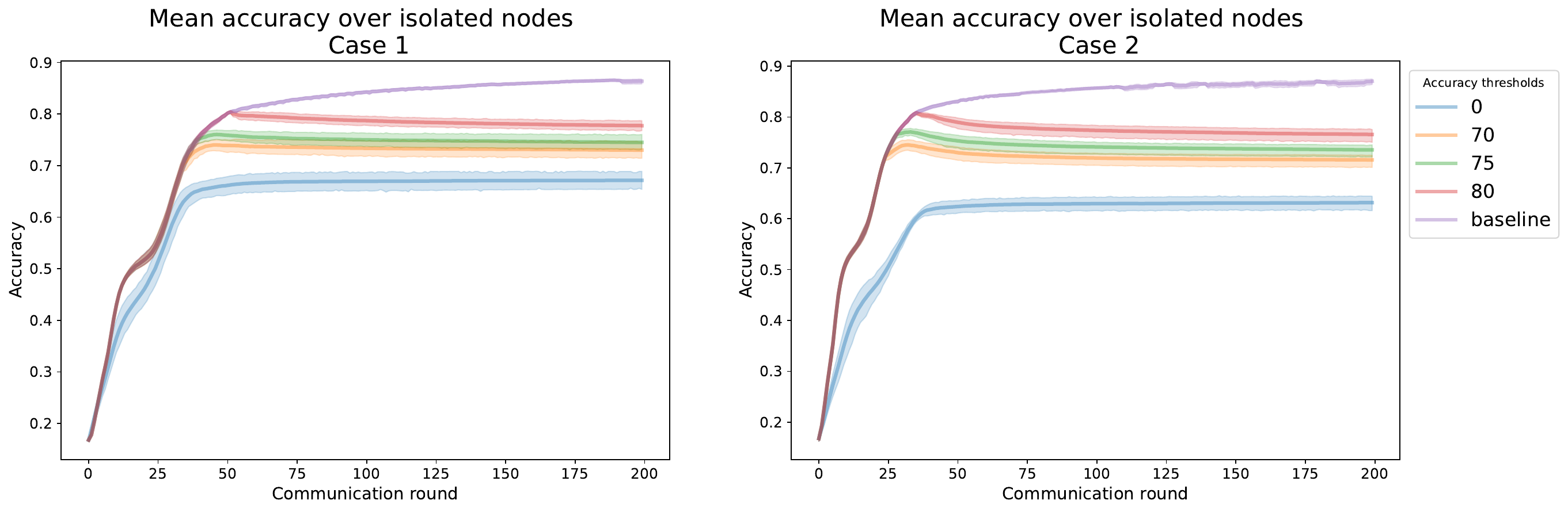}
    \caption{}
    \label{fig:ka_c1c2_persistence_mean_acc_isolated}
    \end{subfigure}
    \begin{subfigure}{\textwidth}
    \centering
    \includegraphics[width=\textwidth]{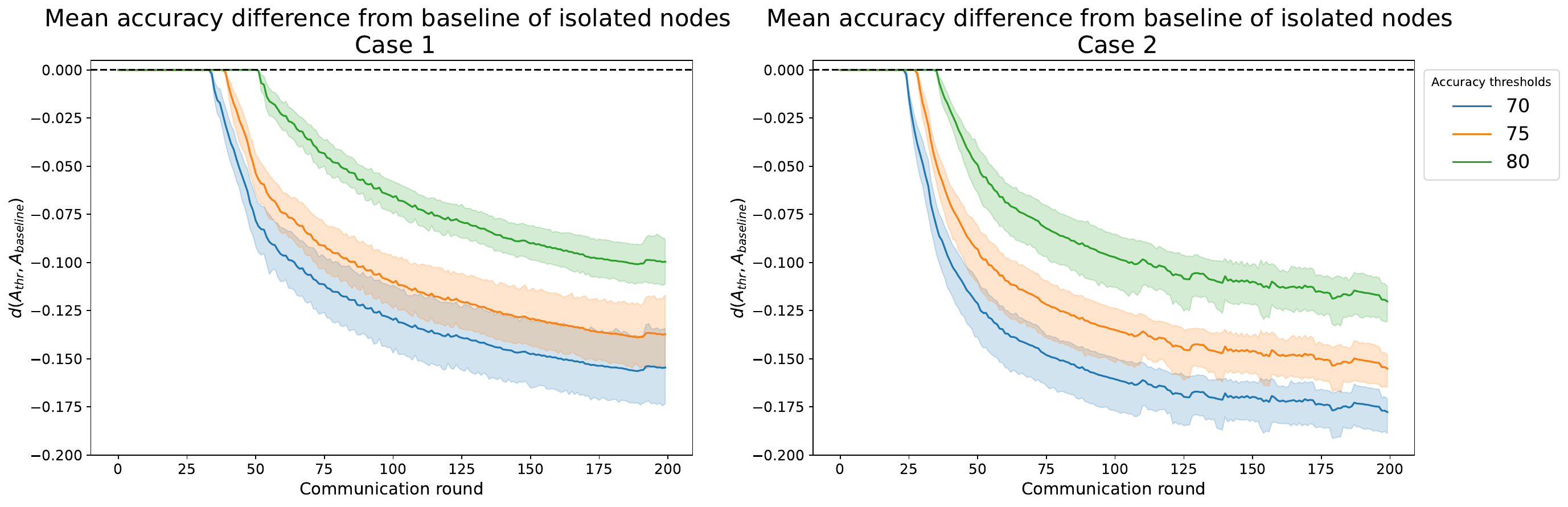}
    \caption{}
    \label{fig:ka_c1c2_persistence_distance_baseline_isolated}
    \end{subfigure}
    \caption{Knowledge persistence analysis (isolated nodes). (a) Mean accuracy over all isolated nodes, for different disruption thresholds. (b) Mean  accuracy difference with respect to baseline over all isolated nodes, for different disruption thresholds.}
    \label{fig:ka_c1c2_persistence_isolated} 
\end{figure}

\begin{figure}[ht]
    \centering
    \begin{subfigure}{\textwidth}
    \centering
    \includegraphics[width=\textwidth]{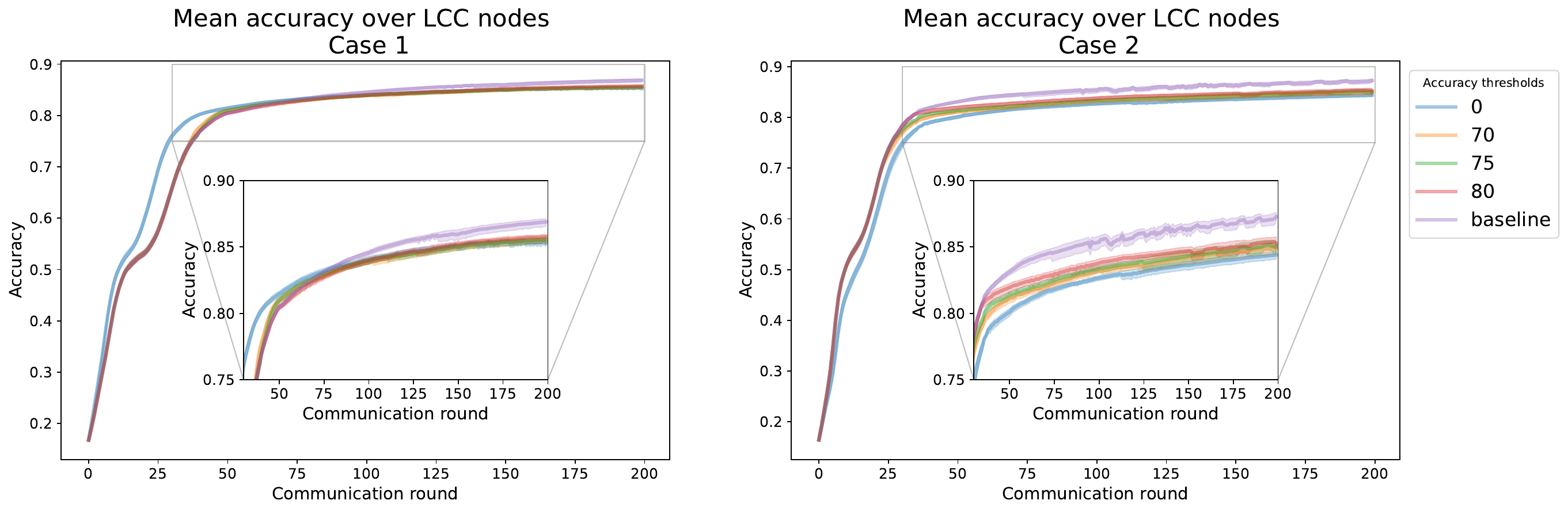}
    \caption{}
    \label{fig:ka_c1c2_persistence_mean_acc_largecluster}
    \end{subfigure}
    \begin{subfigure}{\textwidth}
    \vspace{1em}
    \centering
    \includegraphics[width=\textwidth]{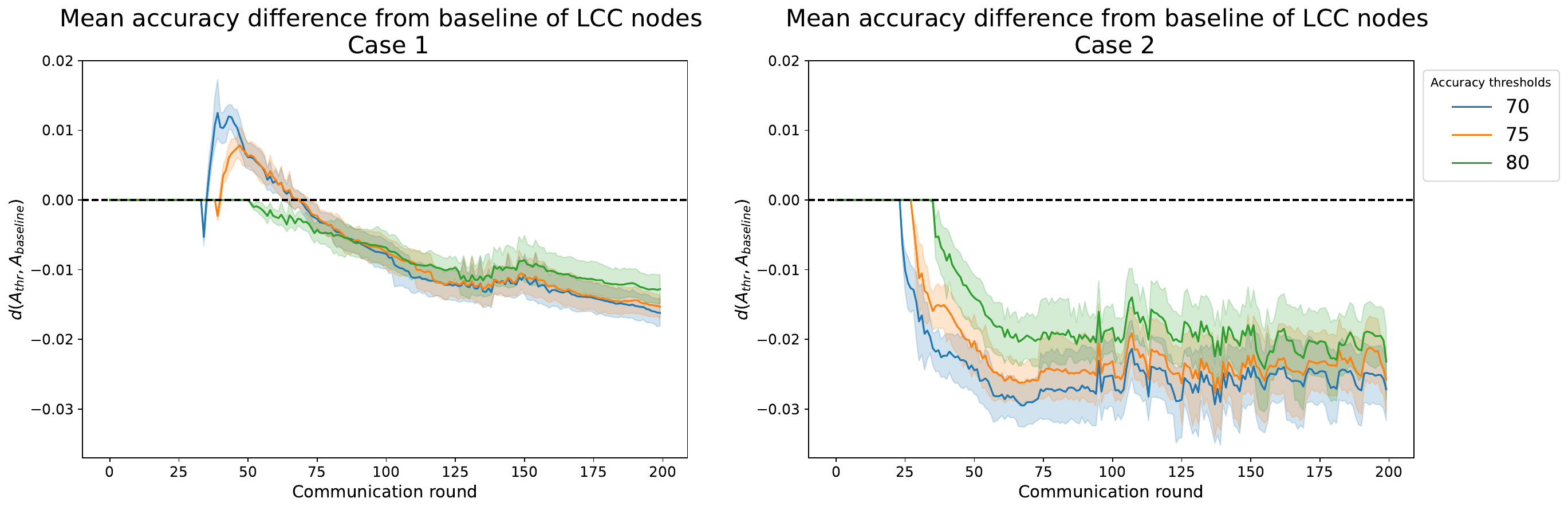}
    \caption{}
\label{fig:ka_c1c2_persistence_distance_baseline_largecluster}
    \end{subfigure}
    
    \caption{Knowledge persistence analysis (largest connected component). (a) Mean accuracy over all nodes in the largest connected component. (b) Mean accuracy difference with respect to baseline over all nodes in the largest connected component.}
    \label{fig:ka_c1c2_persistence_largecluster} 
\end{figure}

As stated in Section~\ref{sec:results}, as a real-world network example, we analyzed the Karate Club Graph~\cite{zachary1977information}. The Karate Club Graph exhibits clear and well-defined community structures due to the actual affiliations and conflicts within the karate club. This makes it highly representative of real community dynamics.
The network is shown in \autoref{fig:ka_net_bef_aft}, before and after the disruption occurs. It seems that the removal of four highly central nodes destroys the community structure. Note that the colouring of nodes is based on the size of the connected component they belong to after disruption.
Following the same considerations presented in Section~\ref{sec:settings_disruption_analysis}, for the Karate Club graph we decided to remove a percentage of nodes that matches the LCC size ratio of the Barabasi-Albert graph. As shown in Figure~\ref{fig:ka_perc}, this amounts to the 12\% which means that we are removing 4 nodes. 
The results obtained confirm the overall system's behavior obtained for the BA network. 
In Figure \ref{fig:ka_accuracy_all}, we
show the accuracy of all the surviving (i.e., non-switched-off) nodes in Case 1 and Case 2 when the disruption happens at different accuracy thresholds.
As we observed for the BA, the later the disruption happens, i.e., going from left to right, the higher
the mean accuracy of the system, resulting in a narrower curve beam. Furthermore, we observe the same clusterization into different groups of similarly performing nodes, based on their connected components' sizes. In Figure \ref{fig:ka_c1c2_persistence_mean_acc_all} we show the average accuracy of the entire system (as for Definition \ref{def: mean_acc_definition}). 
There is a noticeable decrease in accuracy compared to the baseline and the zoomed inset reveals that even though the performance curves are closely clustered, there is a discernible gap between the system’s performance at different accuracy thresholds. This discrepancy becomes more evident when analyzing the percentage differences shown in Figure \ref{fig:ka_c1c2_persistence_distance_baseline_all}. As the accuracy threshold increases, the deviation from the baseline decreases, generally ranging between 4\% and 9\%. Additionally, we observe that Case 2 consistently exhibits slightly lower performance compared to Case 1.
Focusing on the isolated nodes, Figure \ref{fig:ka_c1c2_persistence_isolated} illustrates that the knowledge obtained from the collaborative task persists, resulting in a percentage decrease in accuracy of up to 20\% in the worst case (accuracy threshold of 70\%). Similarly, we observe that the performance of the nodes in the largest connected component (LCC) does not differ significantly from the baseline case (Figure \ref{fig:ka_c1c2_persistence_mean_acc_largecluster} and \ref{fig:ka_c1c2_persistence_distance_baseline_largecluster}). This consistency is due to their continuous stream of connections, which compensates for the disruption.
Overall, the robustness that decentralized learning demonstrated in the Barabási-Albert (BA) network is also evident in the Karate Club Graph. The loss of accuracy compared to the scenario with no disruption is negligible within the LCC, and even for completely isolated nodes, where it remains limited to between 10\% and 20\%, depending on the timing of the disruption. This indicates that decentralized learning maintains a high level of performance despite significant network impairments. In addition, it seems that the removal of four highly central nodes destroys the initial community structure of the Karate Club Graph (see Figure~\ref{fig:ka_net_bef_aft}), so that it does not have an impact on the learning process.


\end{document}